\documentclass[10pt,journal,twoside]{IEEEtran}
\usepackage{graphicx}
\usepackage{amsmath,amssymb} 
\usepackage{color}
\usepackage{times}
\usepackage{epsfig}
\usepackage{threeparttable}
\usepackage{booktabs}
\usepackage{subfigure}
\usepackage{algorithm}
\usepackage{algorithmic}
\usepackage{amsmath}

\usepackage{epsfig}
\usepackage{graphicx}
\usepackage{amsmath}
\usepackage{amssymb}
\usepackage{subfigure}
\usepackage{multirow}
\usepackage{stfloats}
\newtheorem{definition}{Definition}

\ifCLASSOPTIONcompsoc
  \usepackage[nocompress]{cite}
\else
  \usepackage{cite}
\fi

\ifCLASSINFOpdf

\else
 
\fi

\begin{document}
 
\title{Essential Tensor Learning for Multi-view Spectral Clustering}

\author{Jianlong~Wu, 
        Zhouchen~Lin,~\IEEEmembership{Fellow,~IEEE,}
        and~Hongbin~Zha,~\IEEEmembership{Member,~IEEE}
\IEEEcompsocitemizethanks{\IEEEcompsocthanksitem 
Manuscript received July 8, 2018; revised December 14, 2018; accepted May 2, 2019. 
The work of Z.~Lin was supported by 973 Program of China (grant no. 2015CB352502), NSF of China (grant nos. 61625301 and 61731018), Qualcomm, and Microsoft Research Asia. 
The work of H.~Zha was supported by the National Key Research and Development Program of China (grant no. 2017YFB1002601) and National Natural Science Foundation of China (grant nos. 61632003 and 61771026).
The associate editor coordinating the review of this manuscript and approving it for publication was Prof. Sen-ching Samson Cheung. \textit{(Corresponding author: Zhouchen Lin.)}
	
J.~Wu, Z.~Lin, and H.~Zha are with the Key Laboratory of Machine Perception (MOE), School of Electronics Engineering and Computer Science, Peking University, Beijing 100871, China
(e-mail: \{jlwu1992, zlin\}@pku.edu.cn; zha@cis.pku.edu.cn).
}
}

\markboth{IEEE TRANSACTIONS ON IMAGE PROCESSING}
{Wu \MakeLowercase{\textit{et al.}}: Essential Tensor Learning for Multi-view Spectral Clustering}

\IEEEtitleabstractindextext{%
\begin{abstract}
Multi-view clustering attracts much attention recently,
which aims to take advantage of multi-view information to improve the performance of clustering.
However, most recent work mainly focus on self-representation based subspace clustering, which is of high computation complexity. 
In this paper, we focus on the Markov chain based spectral clustering method and propose a novel essential tensor learning method to explore the high order correlations for multi-view representation.
We first construct a tensor based on multi-view transition probability matrices of the Markov chain.
By incorporating the idea from robust principle component analysis, tensor singular value
decomposition (t-SVD) based tensor nuclear norm is imposed to preserve the low-rank property of the essential tensor, which can well capture the principle information from multiple views.
We also employ the tensor rotation operator for this task to better investigate the relationship among views as well as reduce the computation complexity.
The proposed method can be efficiently optimized by the alternating direction  method of multipliers~(ADMM).
Extensive experiments on seven real world datasets corresponding to five different applications show that our method achieves superior performance over other state-of-the-art methods.
\end{abstract}

}

\maketitle

\IEEEdisplaynontitleabstractindextext

\IEEEpeerreviewmaketitle

{\section{Introduction}\label{sec:introduction}}

\IEEEPARstart{C}{lustering} is one of the fundamental tasks in computer vision and pattern recognition, which aims to divide samples into various groups based on their similarity without any prior information. 
It is very useful, especially when the label information is hard to acquire.
There are many clustering based applications, such as image segmentation, dimension reduction, unsupervised classification, etc.
During the past decades, a variety of methods for clustering have been proposed.
Among them, the standard spectral clustering~(SPC)~\cite{ng2002spectral}, 
sparse subspace clustering~(SSC)~\cite{elhamifar2013sparse}, and 
low-rank representation~(LRR)~\cite{liu2013robust} are the most popular methods.
\par
\begin{figure*}[ht]
	\centering
	\includegraphics[width=0.999\linewidth]{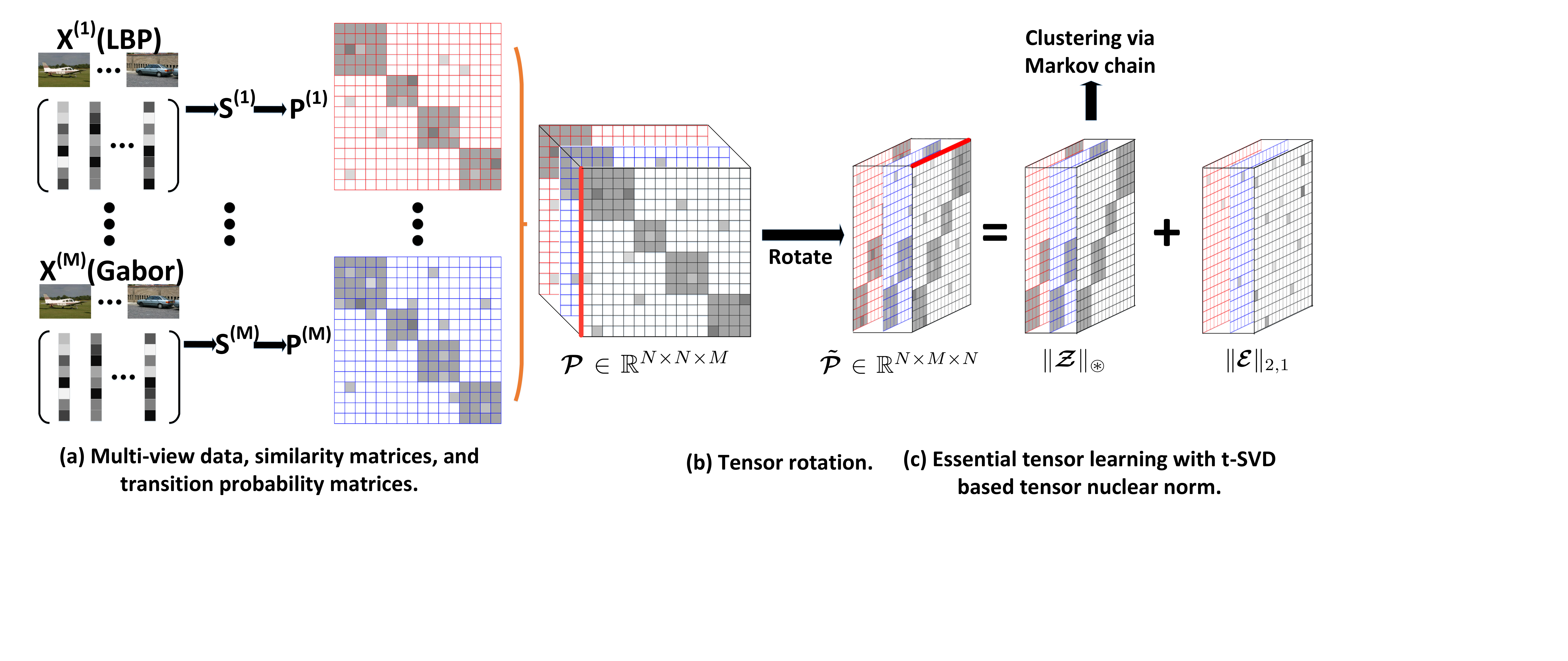}
	\caption{The pipeline of our proposed essential tensor learning for multi-view spectral clustering. 
		For multi-view data $\mathbf{X}^{(i)}(i=1,\cdots,M)$, we first compute the view-specific similarity matrix $\mathbf{S}^{i}\in \mathbb{R}^{N\times N}$ and the corresponding transition probability matrix $\mathbf{P}^{(i)}$ by $\mathbf{P}^{i}=(\mathbf{D}^{(i)})^{-1}\mathbf{S}^{(i)} \in \mathbb{R}^{N\times N}$,
		where $N$ is the total number of samples.
		Then we construct a transition probability matrix tensor $\boldsymbol{\mathcal{P}}$ based on multi-view transition probability matrices. 
		To better explore the high order correlations, we rotate the tensor $\boldsymbol{\mathcal{P}}$ to 
		$\boldsymbol{\mathcal{\tilde{P}}}$~(please pay attention to the rotation of red edge).
		Under the assumption of low-rank and sparse, we learn the essential tensor $\boldsymbol{\mathcal{Z}}$ based on t-SVD based tensor nuclear norm minimization.
		The learned low-rank tensor $\boldsymbol{\mathcal{Z}}$ will
		be used as input to the standard Markov chain method for spectral clustering.}
	\label{fig:overview} 
\end{figure*}  
These single view clustering methods achieve good performance.
In practice, we often acquire data from various domains or feature space.
For example, one object can be described with text, images or videos, 
and different kinds of features can be extracted to represent each of them.
In order to make full use of multi-view information to boost the performance,
many multi-view clustering methods have been derived from these popular single view methods.
\par
Due to the popularity of SSC~\cite{elhamifar2013sparse} and LRR~\cite{liu2013robust}, many self-representation based subspace learning methods~\cite{kumar2011coreg,cao2015diversity,RMSC,wang2017exclusivity,xie2018implicit} are proposed for multi-view clustering.
They achieve promising performances.
But they mainly focus on subspace learning and have high computation complexity.
Another important issue  is that they mainly investigate the correlations from the aspect of pairwise matrices, 
and it is more natural and effective to find comprehensive representation of multi-view from the tensor aspect.
Motivated by the robust multi-view spectral clustering~(RMSC)~\cite{RMSC}, there is a connection between the spectral clustering and Markov chain.
So we mainly focus on the spectral clustering via Markov chain in this paper.
However, RMSC~\cite{RMSC} only learns the shared common information among all views.
While multi-view representations also contain view-specific information,
we hope to explore the high order correlation and find the principle components~\cite{RPCA,TRPCA,RTPCA,zhou2018tensor,kong2018t} of multi-view representations from the tensor aspect based on the Markov chain clustering.
\par
As for tensor decomposition, we not only need to define the rank, but also find a tight convex relaxation of the tensor rank as nuclear norm.
The CANDECOMP/PARAFAC~(CP)~\cite{carroll1970analysis,harshman1970foundations}, Tucker~\cite{tucker1966some} and tensor Singular Value Decomposition~(t-SVD)~\cite{Tensor} are three main tensor decomposition techniques.
However, CP rank is generally NP-hard to compute and its convex relaxation is intractable.
For Tucker decomposition, the commonly used Sum of Nuclear Norms~(SNN)~\cite{huang2014provable} is not a tight convex relaxation of the Tucker rank.
Since t-SVD based tensor nuclear norm has been proven to be the tightest convex relaxation~\cite{zhang2014novel} to $\ell_{1}$-norm of  the tensor multi-rank, so we adopt it. 
With the t-SVD based tensor nuclear norm, our model can well capture both the consistent and view-specific information among multiple views, which will benefit the clustering.
\par
In Fig.~\ref{fig:overview}, we present the framework of our proposed method.
We first construct a similarity matrix and a corresponding transition probability matrix
for features of each view.
Then, we propose to collect these transition probability matrices of multi-view into a 3-order tensor.
In order to better investigate the correlations as well as reduce the computation complexity, we rotate the tensor.
The essential tensor can be learnt via tensor low-rank and sparse decomposition based on tensor nuclear norm minimization defined by the t-SVD.
\par
Main contributions are summarized as follows: \par
\begin{enumerate}
	\item  We propose a novel essential tensor learning method for the Markov chain based spectral clustering.
	With the t-SVD based tensor low-rank constraint and tensor rotation, our method is very effective to learn the principle information for clustering among multiple views. 
	\item  We present an efficient algorithm based on ADMM to solve the proposed problem. 
	\item Our method achieves superior performance compared with the state-of-the-art methods on different datasets for various applications. In the meantime, it also has the lowest computation complexity.
\end{enumerate}

\par
\section{Related Work}

Multi-view clustering has been extensively studied during the past decade.
The standard spectral clustering~(SPC)~\cite{ng2002spectral} is the most classic method, which constructs the similarity matrix first, and then learns the affinity matrix by exploiting the properties of the Laplacian of graph.
Most existing clustering methods are derived from SPC~\cite{ng2002spectral}, and they mainly differ in 
the construction of affinity matrix,
according to which,
existing work can be mainly divided into two classes, including the graph based affinity matrix learning methods and the self-representation based subspace learning methods.
We briefly review some related work.
\par
The graph based methods learn affinity matrix based on the similarity matrix.
For example, \cite{kumar2011cotrain} proposes a co-training approach to search for the clusterings that agree across the views. 
\cite{kumar2011coreg} aims to find the complementary information across views based on a co-regularization method. 
\cite{wang2014multi} tries to find a universal Laplacian embedding for multi-view features using minimax optimization.
The work in \cite{shi2000normalized,zhou2005learning} shows that there is a natural connection between the spectral clustering and the Markov random walk.
Then, \cite{zhou2007spectral} constructs a transition probability matrix of Markov chain on each view, and then combines these matrices via a Markov mixture.
Considering that multi-view data might be noisy, RMSC~\cite{RMSC} 
hopes to recover a shared low-rank transition probability matrix for the Markov chain based spectral clustering.
Recently, \cite{wang2018multiview} proposes the structured low-rank matrix factorization methods for multi-view spectral clustering.
\par
For the second class, multi-view subspace learning methods are derived from the popular SSC~\cite{elhamifar2013sparse} and LRR~\cite{liu2013robust},
which aim to explore the relationships between samples based on self-representation.
Most recent work of multi-view clustering mainly focus on self-representation based subspace learning.
For example, \cite{brbic2018multi} combines the advantages of both LRR and SSC.
\cite{zhang2015low} extends the LRR into multi-view subspace clustering with generalized tensor nuclear norm.
Then \cite{xie2016unifying} adopts the t-SVD based tensor nuclear norm for better representation, and~\cite{yin2018multiview} proposes the tensorial t-product representation.
Zhang et al.~\cite{zhang2017latent} jointly learns the underlying latent representation of features and the multi-view low-rank representation, and then generalize it to combine with deep neural network~\cite{zhang2018generalized}.
To explore the complementary property of multi-view representations,
\cite{cao2015diversity} utilizes the Hilbert Schmidt Independence Criterion~(HSIC) as a diversity term between views, and \cite{wang2017exclusivity} adds an exclusivity term to the structured sparse subspace clustering model~\cite{li2015structured} to preserve the complementary and consistent information.
\par
Besides the above two classes of methods,
there are also some other methods, such as the canonical correlation analysis~(CCA) for multi-view clustering~\cite{chaudhuri2009multi}, multiple kernel learning~\cite{cortes2009learning}, discriminative k-means~\cite{xu2016discriminatively},
and so on.

\section{Notations and Preliminaries}
\subsection{Notations}
For convenience, we summarize the frequently used notations in Table~\ref{tab:notations}. In this paper, we mainly consider the 3-order tensor $\boldsymbol{\mathcal{A}}\in \mathbb{R}^{n_{1}\times n_{2} \times n_{3}}$.
Vector along the $i$-th mode is called the mode-$i$ fiber.
Here, we define the $\ell_{2,1}$-norm of a tensor as the sum of $\ell_{2}$-norm of each mode-$3$ fiber.
$\mathbf{A}_{(i)}$ denote the matricization of $\boldsymbol{\mathcal{A}}$ along the $i$-th mode.
It can be constructed by arranging the mode-$i$ fibers to be the columns of the resulting matrix.
The transpose $\boldsymbol{\mathcal{A}}^{T} \in \mathbb{R}^{n_{2}\times n_{1} \times n_{3}}$ is obtained by transposing each frontal slice and then reversing the order of transposed frontal slices $2$ through $n_3$.
$\boldsymbol{\mathcal{A}}_{f}=\text{fft}(\boldsymbol{\mathcal{A}},[ \ ],3)$ denotes the fast Fourier transformation~(FFT) of a tensor $\boldsymbol{\mathcal{A}}$ along the $3$rd dimension, and we also have $\boldsymbol{\mathcal{A}}=\mathrm{ifft}(\boldsymbol{\mathcal{A}}_{f},[ \ ],3)$.

\par

\begin{table*}[t]
	\centering
	\renewcommand\arraystretch{1.2}
	\caption{Summary of notations in this paper.}
	\label{tab:notations}
	\begin{tabular}{l|l|l|l}
		\hline
		$a$ & A scalar.  & $\mathbf{A}$ & A matrix.  \\
		$\mathbf{a}$  & A vector.  &$\boldsymbol{\mathcal{A}}$  & A tensor.  \\ \hline
		$\Vert \mathbf{A} \Vert_{F}$& $\Vert \mathbf{A} \Vert_{F}=\sqrt{\sum_{ij}A_{ij}^{2}}$.         &  $\Vert \mathbf{A} \Vert_{*}$ & Sum of the singular values.  \\
		$\Vert \mathbf{A} \Vert_{1}$& $\Vert \mathbf{A} \Vert_{1}=\sum_{ij}\vert A_{ij}\vert$. &$\Vert \mathbf{A} \Vert_{2,1}$& $\Vert \mathbf{A} \Vert_{2,1}=\sum_{j}\Vert A(:,j)\Vert_{2}$.  \\ \hline
		$\boldsymbol{\mathcal{A}}_{ijk}$ & The $(i,j,k)$-th entry of $\boldsymbol{\mathcal{A}}$.& $\Vert \boldsymbol{\mathcal{A}} \Vert_{F}$ & $\Vert \boldsymbol{\mathcal{A}} \Vert_{F}=\sqrt{\sum_{ijk}\vert \boldsymbol{\mathcal{A}}_{ijk} \vert^{2}}$.  \\
		$\boldsymbol{\mathcal{A}}(i,:,:)$ &The $i$-th horizontal slice of $\boldsymbol{\mathcal{A}}$.& $\Vert \boldsymbol{\mathcal{A}} \Vert_{1}$ & $\Vert \boldsymbol{\mathcal{A}} \Vert_{1}=\sum_{ijk}\vert \boldsymbol{\mathcal{A}}_{ijk} \vert$. \\
		$\boldsymbol{\mathcal{A}}(:,i,:)$ &The $i$-th lateral slice of $\boldsymbol{\mathcal{A}}$.& $\Vert \boldsymbol{\mathcal{A}} \Vert_{2,1}$ & $\Vert \boldsymbol{\mathcal{A}} \Vert_{2,1} = \sum_{i,j} \Vert \boldsymbol{\mathcal{A}}(i,j,:) \Vert_{2}$. \\
		$\boldsymbol{\mathcal{A}}(:,:,i)$ &The $i$-th frontal slice of $\boldsymbol{\mathcal{A}}$.& $\Vert \boldsymbol{\mathcal{A}} \Vert_{\infty}$ & $\Vert \boldsymbol{\mathcal{A}} \Vert_{\infty}=\max_{ijk}\vert \boldsymbol{\mathcal{A}}_{ijk} \vert$.   \\ 
		$\boldsymbol{\mathcal{A}}_{f}$&$\boldsymbol{\mathcal{A}}_{f}=\text{fft}(\boldsymbol{\mathcal{A}},[ \ ],3)$.& $\Vert \boldsymbol{\mathcal{A}} \Vert_{\circledast}$ & t-SVD based tensor nuclear norm. \\
		$\mathbf{A}^{(i)}$&$\mathbf{A}^{(i)}=\boldsymbol{\mathcal{A}}(:,:,i)$.
		&$\boldsymbol{\mathcal{A}}^{T}$ & The transpose of $\boldsymbol{\mathcal{A}}$.  \\
		$\mathbf{A}_{(i)}$&Mode-$i$ matricization of $\boldsymbol{\mathcal{A}}$.&\\
		\hline
	\end{tabular}
\end{table*}

%
Besides, for a tensor $\boldsymbol{\mathcal{A}}\in \mathbb{R}^{n_{1} \times n_{2} \times n_{3}}$,
we also define the block vectorizing and its inverse operation as $\mathrm{bvec}(\boldsymbol{\mathcal{A}})=[\mathbf{A}^{(1)};\mathbf{A}^{(2)};\cdots;\mathbf{A}^{(n_3)}] \in \mathbb{R}^{n_{1}n_{3}\times n_{2}}$ and $\mathrm{fold}(\mathrm{bvec}(\boldsymbol{\mathcal{A}}))=\boldsymbol{\mathcal{A}}$, respectively.
The block diagonal matrix $\mathrm{bdiag}(\boldsymbol{\mathcal{A}})\in \mathbb{R}^{n_{1}n_{3}\times n_{2}n_{3}} $ and the block circulant matrix $\mathrm{bcirc}(\boldsymbol{{\mathcal{A}}})\in \mathbb{R}^{n_{1}n_{3}\times n_{2}n_{3}}$ are defined by:
\begin{equation}
	\mathrm{bdiag}(\boldsymbol{\mathcal{A}}) :=
	\left[
	\begin{matrix}
		\mathbf{A}^{(1)} & & & \\
		& \mathbf{A}^{(2)}  & & \\
		& & \ddots & \\
		&&& \mathbf{A}^{(n_{3})}
	\end{matrix}
	\right],
	\notag
\end{equation}
		\\
\begin{equation}
	\mathrm{bcirc}(\boldsymbol{{\mathcal{A}}}) :=
	\left[
	\begin{matrix}
		\mathbf{A}^{(1)}    &\mathbf{A}^{(n_{3})}  & \cdots         & \mathbf{A}^{(2)} \\
		\mathbf{A}^{(2)}    &\mathbf{A}^{(1)}      &  \cdots        & \mathbf{A}^{(3)} \\
		\vdots            &\ddots              & \ddots         & \vdots         \\
		\mathbf{A}^{(n_{3})}&\mathbf{A}^{(n_{3}-1)}& \cdots         & \mathbf{A}^{(1)}
	\end{matrix}
	\right].
	\notag
\end{equation}
\subsection{Preliminaries}
To help understand the definition of tensor nuclear norm, we first introduce some related definitions~\cite{Tensor}.
\begin{definition}[\textbf{t-product}]\label{def:t-prod}
	Let $\boldsymbol{{\mathcal{A}}}$ be $n_{1} \times n_{2} \times n_{3}$, and $\boldsymbol{{\mathcal{B}}}$ be $n_{2} \times n_{4} \times n_{3}$. Then the t-product $\boldsymbol{{\mathcal{A}}}*\boldsymbol{{\mathcal{B}}}$ is the $n_{1} \times n_{4} \times n_{3}$ tensor
	\begin{equation}
		\label{fml:t-prod}
		\boldsymbol{{\mathcal{A}}}*\boldsymbol{{\mathcal{B}}} = \mathrm{fold}(\mathrm{bcirc}(\boldsymbol{{\mathcal{A}}})\mathrm{bvec}(\boldsymbol{{\mathcal{B}}})).
	\end{equation}
\end{definition}
\begin{definition}[\textbf{f-diagonal tensor}]
	A tensor is called f-diagonal if each of its frontal slices is diagonal matrix. 
	\label{def:f-diag}
\end{definition}
\begin{definition}[\textbf{Identity tensor}]
	For the identity tensor $\boldsymbol{{\mathcal{I}}} \in \mathbb{R}^{n \times n \times n_{3}}$, its first frontal slice is the identity matrix with size $n \times n$, and all other frontal slices are zero.
	\label{def:t-I}
\end{definition}
\begin{definition}[\textbf{Orthogonal tensor}]
	A tensor $\boldsymbol{{\mathcal{Q}}} \in \mathbb{R}^{n \times n \times n_{3}}$ is orthogonal if it satisfies
	\begin{equation}
		\boldsymbol{{\mathcal{Q}}}^{\mathrm{T}}*\boldsymbol{{\mathcal{Q}}} = \boldsymbol{{\mathcal{Q}}}*\boldsymbol{{\mathcal{Q}}}^{\mathrm{T}} = \boldsymbol{{\mathcal{I}}}.
	\end{equation}
	\label{def:orth-tensor}
\end{definition}
\begin{definition}[\textbf{t-SVD}]
	For a tensor $\boldsymbol{{\mathcal{A}}}\in \mathbb{R}^{n_{1} \times n_{2} \times n_{3}}$, it can be factorized by t-SVD as
	\begin{equation}
		\boldsymbol{{\mathcal{A}}} = \boldsymbol{{\mathcal{U}}} * \boldsymbol{{\mathcal{S}}} * \boldsymbol{{\mathcal{V}}}^{T},
	\end{equation}
	where $\boldsymbol{{\mathcal{U}}} \in \mathbb{R}^{n_{1} \times n_{1} \times n_{3}}$ and $\boldsymbol{{\mathcal{V}}} \in \mathbb{R}^{n_{2} \times n_{2} \times n_{3}}$ are orthogonal, and $\boldsymbol{{\mathcal{S}}} \in \mathbb{R}^{n_{1} \times n_{2} \times n_{3}}$ is f-diagonal.
	\label{def:T-SVD}
\end{definition}
\begin{definition}[\textbf{t-SVD based tensor nuclear norm}]
	The t-SVD based tensor nuclear norm $\Vert \boldsymbol{\mathcal{A}} \Vert_{\circledast}$ of a tensor $\boldsymbol{{\mathcal{A}}}\in \mathbb{R}^{n_{1} \times n_{2} \times n_{3}}$ is defined by the sum of singular values of all the frontal slices of $\boldsymbol{\mathcal{A}}_{f}$:
	\begin{equation}
		\Vert \boldsymbol{\mathcal{A}} \Vert_{\circledast} = \sum_{k=1}^{n_{3}} \Vert \boldsymbol{\mathcal{A}}_{f}^{(k)} \Vert_{*}=\sum_{i=1}^{\min (n_{1}, n_{2})} \sum_{k=1}^{n_{3}} \vert \boldsymbol{\mathcal{S}}_{f}^{(k)}(i,i)\vert,
		\label{eq:tnn}
	\end{equation}
	where $\boldsymbol{\mathcal{S}}_{f}^{(k)}$ is computed by the SVD $\boldsymbol{\mathcal{A}}_{f}^{(k)} = \boldsymbol{{\mathcal{U}}}_{f}^{(k)}  \boldsymbol{{\mathcal{S}}}_{f}^{(k)}  \boldsymbol{{\mathcal{V}}}_{f}^{(k)T}$ of frontal slices of $\boldsymbol{\mathcal{A}}_{f}$.
	\label{def:tnn}
\end{definition}
\begin{figure}[t]
	\centering 
	\includegraphics[width=3.4in]{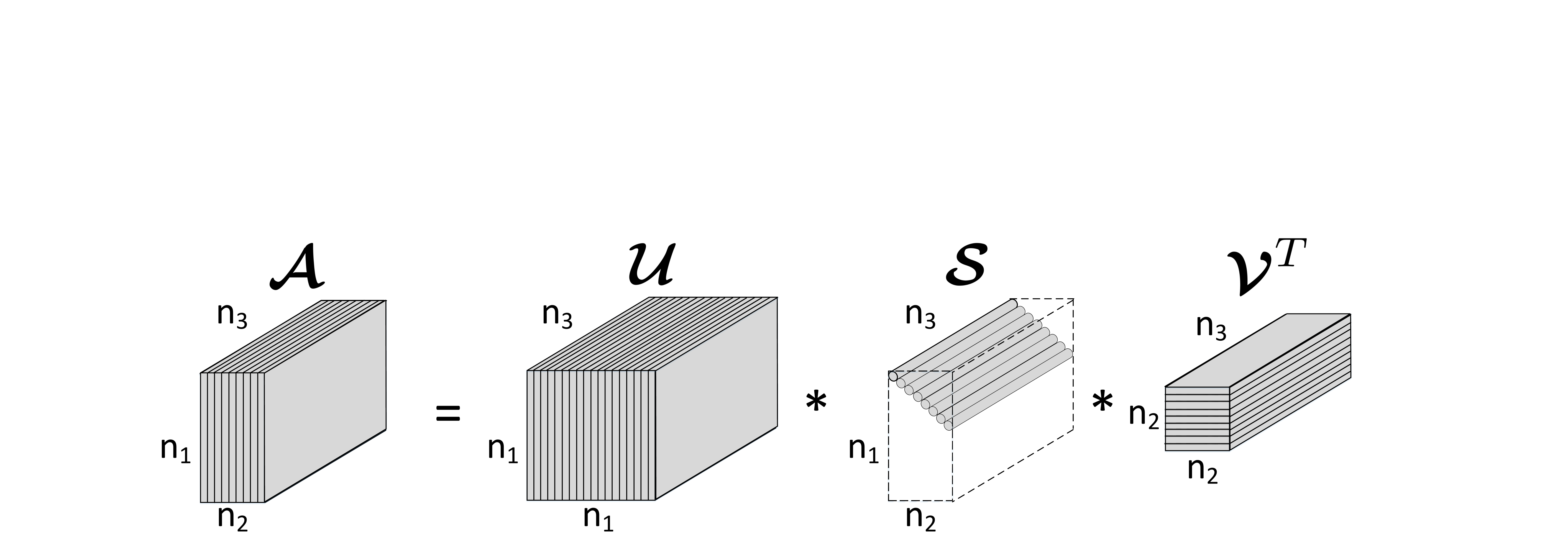}
	\caption{Illustration of the t-SVD decomposition of an $n_{1}\times n_{2} \times n_{3}$ tensor.} 
	\label{fig:t-svd} 
\end{figure}
\section{Essential Tensor Learning for Multi-view Spectral Clustering}
In this section, we first introduce the overview of spectral clustering by Markov chain. Then we present the details and analysis of our proposed essential tensor learning for multi-view spectral clustering~(ETLMSC).
\renewcommand{\algorithmicrequire}{\textbf{Input:}}
\renewcommand{\algorithmicensure}{\textbf{Output:}}
\begin{algorithm}[!tp]
	\caption{Spectral Clustering by Markov Chain}
	\label{al:SCMC} 
	\begin{algorithmic}[1]
		\REQUIRE  Data points $\{ \mathbf{x}_1,\cdots, \mathbf{x}_N \}$.
		\STATE Compute the similarity matrix $\mathbf{S} \in \mathbb{R}^{N\times N}$ with Gaussian kernel $S_{ij}=\exp(-\frac{\Vert \mathbf{x}_i - \mathbf{x}_j \Vert_{2}^{2}}{\sigma^2})$.
		\STATE Construct the weighted graph $\mathbf{G}=(\mathbf{V},\mathbf{E},\mathbf{S})$ and define a random walk over $\mathbf{G}$ with transition probability matrix $\mathbf{P}=\mathbf{D}^{-1}\mathbf{S} \in \mathbb{R}^{N\times N}$ such that it has a unique stationary distribution $\mathbf{\pi}$ satisfying $\mathbf{\pi}=\mathbf{P}^{T}\mathbf{\pi}$.
		\STATE Compute eigenvalues decomposition of the normalized Laplacian matrix $\mathbf{L}^{'}=(\mathbf{\Pi}^{\frac{1}{2}} \mathbf{P} \mathbf{\Pi}^{-\frac{1}{2}}+ \mathbf{\Pi}^{-\frac{1}{2}} \mathbf{P}^{T} \mathbf{\Pi}^{\frac{1}{2}})/2$, where $\mathbf{\Pi}$ is a diagonal matrix with ${\Pi}_{ii}=\mathbf{\pi}(i)$.
		\STATE Adopt the k-means to cluster row vectors of $\mathbf{U} \in \mathbb{R}^{N\times C}$, which consists of $C$ eigenvectors corresponding to the $C$ largest eigenvalues of $\mathbf{L}^{'}$ in the last step, and assign each data point into the corresponding class.
		\ENSURE  Assigned class of each data point.
	\end{algorithmic}
\end{algorithm}
\subsection{Markov Chain based Spectral Clustering}
Denote $\mathbf{X}=[\mathbf{x}_{1}, \cdots, \mathbf{x}_{N}] \in \mathbb{R}^{d\times N}$ as the the matrix of data vectors, where $N$ is the number of data points and $d$ is the dimension of feature vectors.
We first compute the similarity matrix $\mathbf{S}$, where $S_{ij}$ denotes the similarity between data points $\mathbf{x}_{i}$ and $\mathbf{x}_{j}$. Gaussian kernel is commonly used to define their similarity. We have $S_{ij}=\exp(-\frac{\Vert \mathbf{x}_i - \mathbf{x}_j \Vert_{2}^{2}}{\sigma^2})$, where the $\ell_{2}$ distance is adopted and $\sigma$ is the standard deviation. Then we can construct a weighted graph $\mathbf{G}=(\mathbf{V},\mathbf{E},\mathbf{S})$, where the vertices set $\mathbf{V}$ consists of the sample points, the edges set $\mathbf{E}$ denotes the connection between data points, and the similarity $\mathbf{S}$ defines the weight of each edge. 
For spectral clustering~\cite{ng2002spectral}, it tries to find an optimal partition in the weighted graph $\mathbf{G}$.
According to~\cite{shi2000normalized,zhou2005learning},
there is a natural connection between spectral clustering and random walkers on the weighted graph. We first define the transition probability matrix by $\mathbf{P}=\mathbf{D}^{-1}\mathbf{S}$, where $P_{ij}$ denotes the probability of random walk from node $i$ to node $j$, and $\mathbf{D}$ is a diagonal matrix with elements $D_{ii}=\sum_{j}S_{ij}$.
For this Markov chain, we hope the random walk over the graph converges to a
unique and positive stationary distribution $\mathbf{\pi}$, that is $\mathbf{\pi}=\mathbf{P}^{T}\mathbf{\pi}$.
Let $\mathbf{\Pi}$ denote the diagonal matrix with ${\Pi}_{ii}=\pi(i)$, 
then the Laplacian matrix for the Markov chain based spectral clustering can be computed by
$\mathbf{L}=\mathbf{\Pi} - (\mathbf{\Pi} \mathbf{P} + \mathbf{P}^{T} \mathbf{\Pi})/2$.
Denote $C$ as the number of clusters,
the indicator function $\mathbf{f}$ for clustering can be solved by computing the eigenvectors corresponding to the $C$ smallest eigenvalues of the generalized eigenvalue decomposition problem $\mathbf{L}\mathbf{f}=\lambda \mathbf{\Pi} \mathbf{f}$, which is equivalent to the eigenvectors corresponding to the $C$ largest eigenvalues of the normalized Laplacian matrix $\mathbf{L}^{'}=(\mathbf{\Pi}^{\frac{1}{2}} \mathbf{P} \mathbf{\Pi}^{-\frac{1}{2}}+ \mathbf{\Pi}^{-\frac{1}{2}} \mathbf{P}^{T} \mathbf{\Pi}^{\frac{1}{2}})/2$.
Finally, k-means algorithm~\cite{kmeans} is adopted to cluster based on these indicator vectors.
In Algorithm~\ref{al:SCMC}, we briefly summarize the outline for spectral clustering by Markov chains.
For more details, please refer to~\cite{RMSC,zhou2005learning}.
\subsection{The Proposed Method}
Assume that there are $M$ different views in total.
Let $\mathbf{X}^{(i)}=[\mathbf{x}_1^{(i)},\cdots,\mathbf{x}_N^{(i)}]\in \mathbb{R}^{d^{(i)}\times N}$ denote the data matrix of the $i$-th view, where $N$ is the number of samples, $d^{(i)}$ is the dimension of feature vectors in the $i$-th view, and $i$ ranges from $1$ to $M$.
For multi-view spectral clustering via Markov chain, we first compute the similarity matrix $\mathbf{S}^{(i)}\in \mathbb{R}^{N\times N}$, construct the weighted graph $G^{(i)}$,
and compute the transition probability matrix $\mathbf{P}^{(i)}$ for each view.
According to Algorithm~\ref{al:SCMC}, we can see that the transition probability matrix $\mathbf{P}$ plays a very important role in the clustering by Markov chain.
So we mainly focus on how to learn an essential transition probability matrix for spectral clustering based on the multi-view $\mathbf{P}^{(i)}, i=1, \cdots, M$. 
\par
RMSC~\cite{RMSC} hopes to capture the shared information among multi-view transition probability matrices.
It divides each $\mathbf{P}^{i}$ into two parts: a shared probability matrix $\mathbf{Z}$ describing important information for clustering, and view-specific deviation error matrix $\mathbf{E}^{(i)}$.
As the number of clusters is much smaller than the sample number, RMSC imposes low-rank constraint on $\mathbf{Z}$. It also assumes that the error matrix should be sparse. Then the objective function for RMSC~\cite{RMSC} is formulated as
\begin{equation}
	\min_{\mathbf{Z},\mathbf{E}^{(i)}} \Vert\mathbf{Z}\Vert_{*}  +  \lambda 
	\sum_{i=1}^{M} \Vert \mathbf{E}^{(i)} \Vert_{1} \
	s.t. \ \mathbf{P}^{(i)}  =  \mathbf{Z}  +  \mathbf{E}^{(i)}, i = 1,\cdots, M,
	\label{eq:rmsc}
\end{equation}
where $\lambda$ is a balance parameter.
\par
RMSC only learns the shared common information among multiple views.
However, each view also contains unique information that is useful for clustering.
Motivated by this, we hope to explore high order correlations among multiple views based on tensor representation.
\par
We divide each $\mathbf{P}^{i}$ into two parts $\mathbf{P}^{(i)}=\mathbf{Z}^{(i)}+ \mathbf{E}^{(i)}$. 
Then we construct a 3-order tensor $\boldsymbol{\mathcal{Z}}$ by collecting all $\mathbf{Z}^{(i)}$.
As multi-view features are extracted from the same objects, different $\mathbf{Z}^{(i)}$ also contains some similar information. 
In the meantime, the number of clusters is much smaller than the sample number.
So the tensor $\boldsymbol{\mathcal{Z}}$ should be low-rank.
We use the t-SVD based tensor nuclear norm $\Vert \cdot \Vert_{\circledast}$ to regularize $\boldsymbol{\mathcal{Z}}$ and get the primary objective function for our model:
\begin{equation}
	\min_{\mathbf{Z},\mathbf{E}^{(i)}}  \!  \Vert  \boldsymbol{\mathcal{Z}}  \Vert_{\circledast}  \! +  \!    \lambda 
	\sum_{i=1}^{M}  \Vert  \mathbf{E}^{(i)}   \Vert_{1}   \
	s.t. \ \mathbf{P}^{(i)}  \!  =   \!  \mathbf{Z}^{(i)}   \!  +  \!   \mathbf{E}^{(i)}, i  =  1,\cdots, M.
	\label{eq:model1}
\end{equation}
The minimization of low-rank tensor can help us find the essential information among different views.
Specifically, the consistent information among multiple views may be represented by several principle components of the t-SVD, and view-specific information can be preserved in other singular values of the corresponding slice of the f-diagonal tensor  $\boldsymbol{\mathcal{S}}	$, which is computed by the t-SVD..
By constructing a 3-order transition probability tensor $\boldsymbol{\mathcal{P}}\in \mathbb{R}^{N\times N \times M}$, where $\mathbf{P}^{(i)}$ is the $i$-th frontal slice of the tensor $\boldsymbol{\mathcal{P}}$, the above problem can be reformulated as the tensor form:

\begin{equation}
	\min_{\boldsymbol{\mathcal{Z}},\boldsymbol{\mathcal{E}}} \Vert \boldsymbol{\mathcal{Z}} \Vert_{\circledast} + \lambda \Vert \boldsymbol{\mathcal{E}} \Vert_{1}, \  s.t. \ \boldsymbol{\mathcal{P}}=\boldsymbol{\mathcal{Z}}+\boldsymbol{\mathcal{E}}.
	\label{eq:pri_obj}
\end{equation}

\par
Instead of optimizing the above problem, we first rotate the original transition probability tensor $\boldsymbol{\mathcal{P}} \in \mathbb{R}^{N\times N \times M}$ into $\boldsymbol{\mathcal{\tilde{P}}} \in \mathbb{R}^{N\times M \times N}$, which can be seen in the middle part of Fig.~\ref{fig:overview}~(please pay attention to the rotation of the red edge of the tensor).
This tensor rotation can be easily achieved by the $shiftdim$ function in Matlab.
There are mainly two advantages for this operation.
First, according to the definition of t-SVD, FFT operates along the third dimension of the tensor and then we perform SVD in each frontal slice.
As we hope to capture the essential information among all views, SVD in each slice with the information of multi-view and all samples is more meaningful. 
Moreover, FFT along the feature dimension can preserve the relationship among views.
Second, this rotation can largely reduce the computation complexity in optimization, which will be analysed in the subsection~\ref{sec:4.4}.
\par
Besides, for the error term, if one sample contains much noise and outliers, transition probability vectors in the tensor related to this sample will be influenced. Noises in these vectors are not sparse, so $\ell_{2}$-norm regularization on vectors is more proper.
As noisy samples should be sparse, tensor $\ell_{2,1}$-norm works.
It is more robust to outliers and noises.
So we use $\ell_{2,1}$-norm to characterize the sparsity property.
Then the final objective function of our proposed ETLMSC method can be reformulated as follows:
\begin{equation}
	\min_{\boldsymbol{\mathcal{Z}},\boldsymbol{\mathcal{E}}} \Vert \boldsymbol{\mathcal{Z}} \Vert_{\circledast} + \lambda \Vert \boldsymbol{\mathcal{E}} \Vert_{2,1}, \  s.t. \ \boldsymbol{\mathcal{\tilde{P}}}=\boldsymbol{\mathcal{Z}}+\boldsymbol{\mathcal{E}},
	\label{eq:final_obj}
\end{equation}
where $\boldsymbol{\mathcal{\tilde{P}}}$ denotes the rotated transition probability tensor.
For the tensor $\boldsymbol{\mathcal{E}}$ after rotation, the 
$\ell_{2,1}$-norm is defined as the sum of $\ell_{2}$-norm of each fiber along the coefficient dimension.
According to the definition of $\ell_{2,1}$-norm and matricization in Table~\ref{tab:notations}, we have $\Vert \boldsymbol{\mathcal{E}} \Vert_{2,1}=\Vert \mathbf{E}_{(3)}\Vert_{2,1}$, which is helpful to the optimization of $\boldsymbol{\mathcal{E}}$.
\renewcommand{\algorithmicrequire}{\textbf{Input:}}
\renewcommand{\algorithmicensure}{\textbf{Output:}}
\begin{algorithm}[!tp]
	\caption{Essential Transition Probability Tensor Learning for Multi-view Spectral Clustering}
	\label{al:RTRPCA} 
	\begin{algorithmic}[1]
		\REQUIRE  Multi-view data $\mathbf{X}^{(i)}\in \mathbb{R}^{d_i\times N},\ i=1,2,\cdots, M$.
		\STATE Compute $\mathbf{S}^{i}$ and $\mathbf{P}^{i}$, and construct the rotated tensor 
		$\boldsymbol{\mathcal{\tilde{P}}}$.
		Set $k=0$, $\boldsymbol{\mathcal{{L}}}^{0}=\boldsymbol{\mathcal{{E}}}^{0}=\boldsymbol{{\mathcal{Y}}}^{0}=\mathbf{0}$, $\mu^{0}=10^{-3}$, $\rho=2$, $\mu^{max}=10^{8}$, and $\epsilon=10^{-6}$.
		\WHILE{ not converged}
		\STATE Fix $\boldsymbol{\mathcal{E}}^{k}$. Update $\boldsymbol{\mathcal{Z}}^{k+1}$ by Eq.~\ref{eq:tubal_sh}.
		\STATE Fix $\boldsymbol{\mathcal{Z}}^{k+1}$. Update $\boldsymbol{\mathcal{E}}^{k+1}$ by Eq.~\ref{eq:te_obj}.
		\STATE Update $\boldsymbol{\mathcal{Y}}^{k+1}$ by Eq.~\ref{eq:y_upd}.
		\STATE $\mu^{k+1}=\min(\rho \mu^{k}, \mu^{max})$.
		\STATE Check the convergence conditions: \\
		$\Vert \boldsymbol{\mathcal{Z}}^{k+1} - \boldsymbol{\mathcal{Z}}^{k}\Vert_{\infty}\leq \epsilon$, 
		$\Vert \boldsymbol{\mathcal{E}}^{k+1} - \boldsymbol{\mathcal{E}}^{k}\Vert_{\infty}\leq \epsilon$, \\
		$\Vert \boldsymbol{\mathcal{\tilde{P}}}-\boldsymbol{\mathcal{Z}}^{k+1} - \boldsymbol{\mathcal{E}}^{k+1}\Vert_{\infty}\leq \epsilon$.
		\STATE $k=k+1$.
		\ENDWHILE
		\ENSURE $\boldsymbol{\mathcal{Z}}^{k+1}$ and $\boldsymbol{\mathcal{E}}^{k+1}$.
	\end{algorithmic}
\end{algorithm}
\subsection{Optimization}
We adopt the alternating direction method of multipliers~(ADMM)~\cite{LADM} to solve Eq.~(\ref{eq:final_obj}).
The augmented Lagrangian function can be formulated as follows:
\begin{align}
& \mathcal{L}(\boldsymbol{\mathcal{Z}},\boldsymbol{\mathcal{E}})= \Vert \boldsymbol{\mathcal{Z}} \Vert_{\circledast} + \lambda\Vert \boldsymbol{\mathcal{E}} \Vert_{2,1} \notag \\
& + \langle \boldsymbol{\mathcal{Y}}, \boldsymbol{\mathcal{\tilde{P}}}-\boldsymbol{\mathcal{Z}}-\boldsymbol{\mathcal{E}} \rangle +
\frac{\mu^{k}}{2} \Vert \boldsymbol{\mathcal{\tilde{P}}}-\boldsymbol{\mathcal{Z}}-\boldsymbol{\mathcal{E}} \Vert_{F}^{2} \notag \\
& = \Vert \boldsymbol{\mathcal{Z}} \Vert_{\circledast} \! + \! \lambda\Vert \boldsymbol{\mathcal{E}} \Vert_{2,1} 
\! + \! \frac{\mu^{k}}{2} \Vert \boldsymbol{\mathcal{\tilde{P}}}-\boldsymbol{\mathcal{Z}}-\boldsymbol{\mathcal{E}} + \boldsymbol{\mathcal{Y}}/\mu^{k} \Vert_{F}^{2} ,
\label{eq:lagrangian}
\end{align}
where $\mu^{k}> 0$ is a penalty parameter at $k$-th iteration and $\boldsymbol{\mathcal{Y}}$ is a Lagrange multiplier.
ADMM alternately updates each variable as follows.
\par
$\boldsymbol{\mathcal{Z}}$ sub-problem:
\begin{align}
	\boldsymbol{\mathcal{Z}}^{k+1}  =  \arg \min_{\boldsymbol{\mathcal{Z}}} \Vert \boldsymbol{\mathcal{Z}} \Vert_{\circledast} 
	+  \frac{\mu^{k}}{2} \Vert \boldsymbol{\mathcal{Z}}  -  \left( \boldsymbol{\mathcal{\tilde{P}}}  -  \boldsymbol{\mathcal{E}}^{k}  +  \boldsymbol{\mathcal{Y}}^{k}/\mu^{k} \right) \Vert_{F}^{2},
	\label{eq:z_opt}
\end{align}
which is a t-SVD based tensor nuclear norm minimization problem. 
According to~\cite{TWIST}, it has the following close-form solution with the tensor tubal-shrinkage operator:
\begin{equation}
	\boldsymbol{\mathcal{Z}}^{k+1}= \mathcal{C}_{\mu'}(\boldsymbol{\mathcal{\tilde{P}}}-\boldsymbol{\mathcal{E}}^{k} + \boldsymbol{\mathcal{Y}}^{k}/\mu^{k})
	=\boldsymbol{\mathcal{U}} \ast \mathcal{C}_{\mu'}(\boldsymbol{\mathcal{S}}) \ast \boldsymbol{\mathcal{V}}^{T},
	\label{eq:tubal_sh}
\end{equation}
where $\mu'=N\cdot \mu^{k}$, $\boldsymbol{\mathcal{\tilde{P}}}-\boldsymbol{\mathcal{E}}^{k} + \boldsymbol{\mathcal{Y}}^{k}/\mu^{k} =\boldsymbol{\mathcal{U}} \ast \boldsymbol{\mathcal{S}} \ast \boldsymbol{\mathcal{V}^{T}}$ and $\mathcal{C}_{\mu'}(\boldsymbol{\mathcal{S}}) = \boldsymbol{\mathcal{S}}*\boldsymbol{\mathcal{J}}$. $\boldsymbol{\mathcal{J}}\in \mathbb{R}^{N\times M \times N}$ is an f-diagonal tensor whose diagonal element in the Fourier domain is $\boldsymbol{\mathcal{J}}_{f}(i,i,j) = \max (1 - \frac{\mu'}{{ \boldsymbol{\mathcal{S}}}_{f}^{(j)}(i,i)},0)$.

\par
$\boldsymbol{\mathcal{E}}$ sub-problem:
\begin{equation}
	\boldsymbol{\mathcal{E}}^{k+1}  =  \arg \min_{\boldsymbol{\mathcal{{E}}}} \lambda\Vert  \boldsymbol{\mathcal{E}} \Vert_{2,1} 
	+  \frac{\mu^{k}}{2} \Vert \boldsymbol{\mathcal{E}}  -  \left( \boldsymbol{\mathcal{\tilde{P}}}  -  \boldsymbol{\mathcal{Z}}^{k+1}  +  \boldsymbol{\mathcal{Y}}^{k}/\mu^{k} \right)  \Vert_{F}^{2}.
	\label{eq:te_obj}
\end{equation}
As the $\ell_{2,1}$-norm of the tensor $\boldsymbol{\mathcal{E}}$ is defined as the sum of $\ell_{2}$-norm of each mode-$3$ fiber,
we matricize each tensor along the $3$rd mode. 
So we have $\Vert \mathbf{E}_{(3)}^{k+1} \Vert_{2,1} = \Vert \boldsymbol{\mathcal{E}}^{k+1} \Vert_{2,1}$.
It can be transformed into the matrix form:
\begin{equation}
\begin{aligned}
\mathbf{E}_{(3)}^{k+1} & =  \arg \min_{\mathbf{E}_{(3)}} \lambda\Vert \mathbf{E}_{(3)} \Vert_{2,1}  \\
&+ \frac{\mu^{k}}{2} \Vert \mathbf{E}_{(3)} - \left( \mathbf{\tilde{P}}_{(3)}-\mathbf{Z}^{k+1}_{(3)} + \mathbf{Y}^{k}_{(3)}/\mu^{k} \right) \Vert_{F}^{2}.
\label{eq:me_obj}
\end{aligned}
\end{equation}

Let $\mathbf{D}=\mathbf{\tilde{P}}_{(3)}-\mathbf{Z}^{k+1}_{(3)} + \mathbf{Y}^{k}_{(3)}/\mu^{k}$,
and according to~\cite{liu2013robust}, the problem in Eq.~(\ref{eq:me_obj}) has the following close-form solution:
\begin{equation}\label{E-subproblem-solution}
	\mathbf{E}_{(3):,i}^{k+1} =
	\left\{
	\begin{aligned}
		&\frac{||\mathbf{D}_{:,i}||_2 - \frac{\lambda}{\mu^{k}}}{||\mathbf{D}_{:,i}||_2}\mathbf{D}_{:,i}, \ \  \text{if} \ \ ||\mathbf{D}_{:,i}||_2 >  \frac{\lambda}{\mu^{k}}\\
		&\mathbf{0}, \qquad \qquad \qquad \quad  \text{otherwise.} \\
	\end{aligned}
	\right.
\end{equation}
where $\mathbf{D}_{:,i}$ represents the $i$-th column of the matrix $\mathbf{D}$. After we get $\mathbf{E}_{(3)}^{k+1}$, we transform it into the tensor form.
\par
Update multipliers:
\begin{equation}
	\boldsymbol{\mathcal{Y}}^{k+1}  = \boldsymbol{\mathcal{Y}}^{k} + \mu^{k} \left( \boldsymbol{\mathcal{\tilde{P}}} - \boldsymbol{\mathcal{Z}}^{k+1} - \boldsymbol{\mathcal{E}}^{k+1} \right).
	\label{eq:y_upd}
\end{equation}
The whole optimization process is summarized in Algorithm~\ref{al:RTRPCA}.
After we learn the essential
transition probability tensor $\boldsymbol{\mathcal{Z}}\in \mathbb{R}^{N\times M \times N}$, we compute the essential transition probability matrix $\mathbf{Z}^{*} \in \mathbb{R}^{N\times N}$ by summing its lateral slices as
$\mathbf{Z}^{*}=\sum_{i=1}^{M} \boldsymbol{\mathcal{Z}}(:,i,:)$. Then we put $\mathbf{Z}^{*}$ into the second step of Algorithm~\ref{al:SCMC} to replace the transition probability matrix $\mathbf{P}$, and we can get the final clustering result.
\subsection{Convergence and Complexity}~\label{sec:4.4}
At each iteration, we can get the close-form solution of $\boldsymbol{\mathcal{Z}}^{k+1}$ and $\boldsymbol{\mathcal{E}}^{k+1}$.
In~\cite{LADM}, the convergence of ADMM with two blocks of variables has already been proved.
Accordingly, our algorithm will converge to an optimal solution.
\par
For the computation complexity, at each iteration, it takes $\mathcal{O}(MN^{2})$ to compute the close-form solution of $\boldsymbol{\mathcal{E}}$.
As for updating $\boldsymbol{\mathcal{Z}}$, on the one hand, we need to calculate the FFT and inverse FFT of a $N\times M \times N$ tensor along the third dimension, which takes $\mathcal{O}(MN^{2}\log(N))$. On the other hand, in the Fourier domain, we need to compute the SVD of each frontal slice of a tensor with size $N\times M \times N$, which takes $\mathcal{O}(M^{2}N^{2})$. So we need $\mathcal{O}(M^{2}N^{2}+MN^{2}\log(N))$ in total to compute the close-form solution of $\boldsymbol{\mathcal{Z}}$ under tensor rotation operation. However, if we do not rotate the tensor, we need $\mathcal{O}(MN^{3}+MN^{2}\log(M))$. As the number of views $M$ is much smaller than the number of samples $N$ in multi-view setting, that is $M \ll N$ and $M \leq \log(N)$. Therefore, we can see that the computation complexity is largely reduced by the tensor rotation.
Denote $K$ as the number of iterations, the complexity to learn the essential tensor in Algorithm~\ref{al:RTRPCA} is $\mathcal{O}(KMN^{2}(M+\log(N)))$, which is relatively efficient.
\par
After we get the essential transition probability matrix, we adopt the Markov chain based spectral clustering to get the final result, which usually cost $\mathcal{O}(N^{3})$. Therefore, the overall complexity is $\mathcal{O}(N^{3}+KMN^{2}(M+\log(N)))$.
\begin{figure*}[t]
	\centering 
	\includegraphics[width=7.1in]{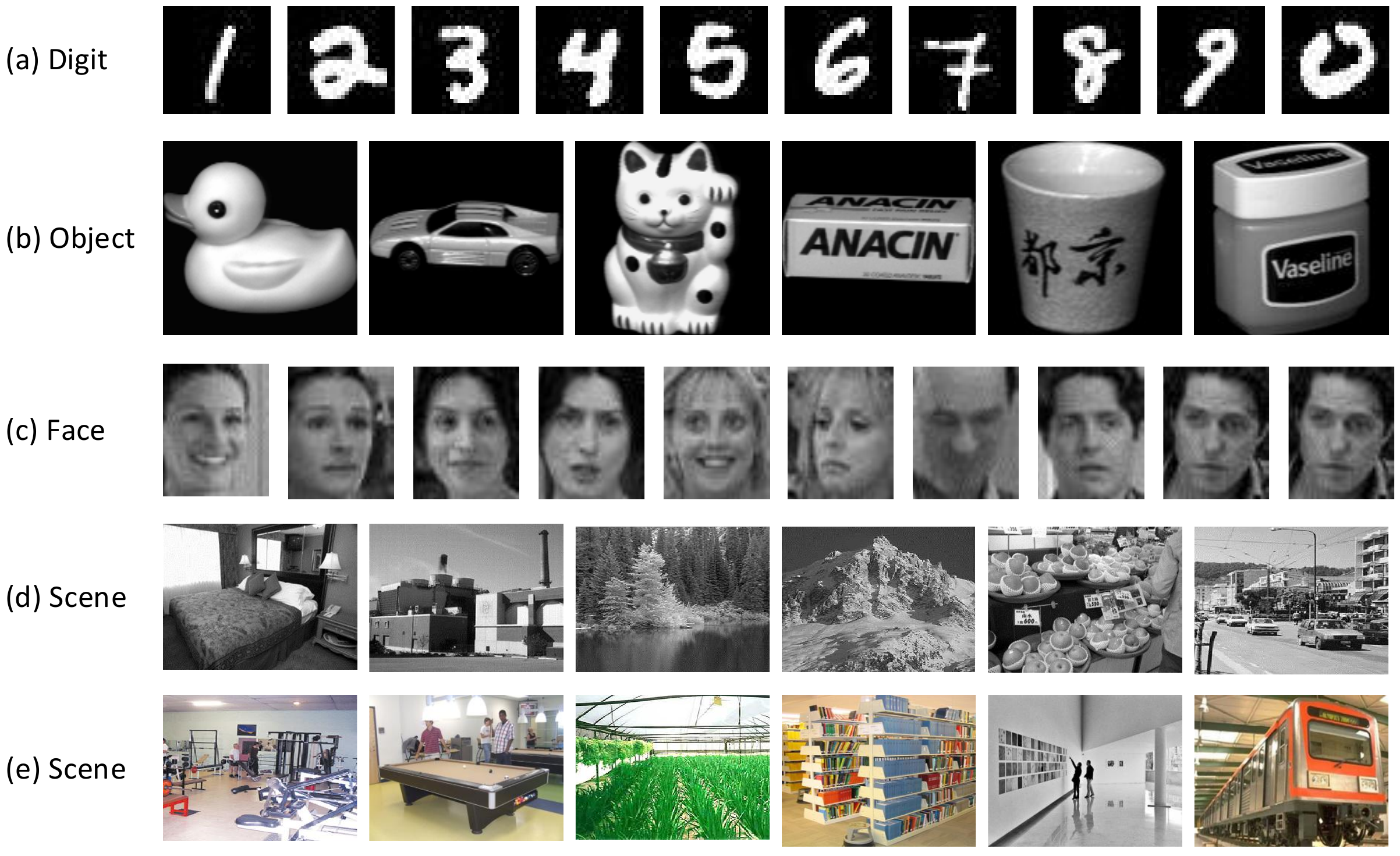}
	\caption{Some sample images of these image datasets for various applications. (a) The UCI-Digits dataset; (b) The COIL-20 dataset; (c) The Notting-Hill dataset; (d) The Scene-15 dataset; (e) The MITIndoor-67 dataset.} 
	\label{fig:datasets_sam} 
\end{figure*}
\section{Experiments}
\subsection{Experimental Settings}
\subsubsection{Datasets}
We adopt seven commonly used real world datasets, which cover five different applications, including news article clustering, digit clustering, generic object clustering, face clustering, and scene clustering.
In Table~\ref{dataset}, we summarize the statistic information of these seven datasets.
Some samples of these image datasets are presented in Fig~\ref{fig:datasets_sam}.
 We briefly introduce these datasets as follows. \\
\textbf{BBC-Sport}~\cite{greene06icml}~\footnote{http://mlg.ucd.ie/datasets} contains $737$ documents from
the BBC Sport website corresponding to sports news in five
topical areas, including the athletics, cricket, football, rugby, and tennis.
There are two different views in total.
\\
\textbf{UCI-Digits}~\cite{UCI} consists of $2,000$ digits images corresponding to $10$ classes.
Same to~\cite{RMSC},
we extract three different features to represent these digit images, including Fourier coefficients, pixel averages and morphological features.
\\
\begin{table}[!t]
	\small
	\renewcommand{\arraystretch}{1.2}
	\begin{centering}
		\begin{threeparttable}[]
			\caption{Statistics of different datasets.}\label{dataset}
			\begin{minipage}{12cm}
				\begin{tabular}{@{}lcccc}
					\toprule[0.4pt] 
					Dataset         &Images &Objective &Clusters &Views  \\ \hline
					BBC-Sport               &773  &Text &5 & 2 \\
					UCI-Digits    &2000  &Digit &10 & 3\\
					COIL-20           &1440 &Object &20 & 3\\ 
					Notting-Hill        &4660 &Video Face &5 & 3\\ 
					Scene-15    &4485  &Scene &15 & 3\\
					MITIndoor-67        &5360 &Scene &67 & 4\\  
					Caltech-101        &8677 &Object &101 & 4\\  
					\bottomrule[0.4pt]
				\end{tabular}
			\end{minipage}
		\end{threeparttable}
	\end{centering}
\end{table}
\textbf{COIL-20}~\footnote{http://www.cs.columbia.edu/CAVE/software/softlib/} is the abbreviation of the Columbia object image library dataset, which contains $1,440$ images of $20$ object categories. Each category contains 72 images and all images are normalized to size $32 \times 32$.
For this datasets, we also extract three types of features~(intensity, LBP~\cite{LBP} and Gabor~\cite{gabor} features), which is same to~\cite{zhang2015low,xie2016unifying}.
\\
\textbf{Notting-Hill}~\cite{zhang2009character} is a video based face dataset, which is collected from the movie ``Notting-Hill''. It contains $4,660$ faces of $5$ main casts in $76$ tracks. 
All face images are with size $50\times 40$.
Intensity, LBP~\cite{LBP} and Gabor~\cite{gabor} features are extracted for representation.\\
\textbf{Scene-15}~\cite{fei2005bayesian} has $15$  natural scene categories with both indoor and outdoor environments, including industrial, store, bedroom, kitchen, and etc. There are $4,485$ images in total. Similar to~\cite{xie2016unifying}, we extract three kinds of image features for representation, including PHOW~\cite{PHOW}, LBP~\cite{LBP}, and CENTRIST~\cite{wu2011centrist}.
\\
\textbf{MITIndoor-67}~\cite{quattoni2009recognizing} contains $15K$ indoor images of $67$ categories. Same to~\cite{xie2016unifying}, the training subset which has $5,360$ images is adopted for clustering.
Besides the three kinds of features for  Scene-15, we also extract deep features based on pretrained VGG-VD~\cite{simonyan2014very} network to improve the performance.
\\
\textbf{Caltech-101}~\cite{fei2007learning} includes $8,677$ object images of $101$ categories. For each category, it has about $40$ to $800$ images. This dataset is the largest dataset used in all these related multi-view clustering methods.
We adopt all these images of $101$ classes to test the performance of clustering, which is same to~\cite{xie2016unifying}.
Besides the three kinds of features for  Scene-15, 
the Inception V3~\cite{szegedy2016rethinking} network is used to extract deep features.
\begin{table*}[t]
	\centering
	\renewcommand\arraystretch{1.2}
	\caption{Experimental results on the BBC-Sport and the UCI-Digit datasets. For ETLMSC, we set $\lambda=0.03$ and $\lambda=0.007$ for these two datasets, respectively.}
	\label{tab:bbc-uci}
	\begin{tabular}{ccccccccccccc}
		\hline
		\multicolumn{1}{|c|}{Datasets} & \multicolumn{6}{c|}{BBC-Sport} & \multicolumn{6}{c|}{UCI-Digits}  \\ \hline \hline
		\multicolumn{1}{|c|}{Methods} & \multicolumn{1}{c|}{NMI} & \multicolumn{1}{c|}{ACC} & \multicolumn{1}{c|}{AR} & \multicolumn{1}{c|}{F-score} & \multicolumn{1}{c|}{Precision} & \multicolumn{1}{c|}{Recall} & \multicolumn{1}{c|}{NMI} & \multicolumn{1}{c|}{ACC} & \multicolumn{1}{c|}{AR} & \multicolumn{1}{c|}{F-score} & \multicolumn{1}{c|}{Precision} & \multicolumn{1}{c|}{Recall} \\ \hline
		\multicolumn{1}{|c|}{SPC$_{best}$}        & \multicolumn{1}{c|}{0.735}    & \multicolumn{1}{c|}{0.853 }    & \multicolumn{1}{c|}{0.744}   & \multicolumn{1}{c|}{0.798}        & \multicolumn{1}{c|}{0.804}          & \multicolumn{1}{c|}{0.792}       
		& \multicolumn{1}{c|}{0.642}    & \multicolumn{1}{c|}{0.731 }    & \multicolumn{1}{c|}{0.545}   & \multicolumn{1}{c|}{0.591}        & \multicolumn{1}{c|}{0.582}          & \multicolumn{1}{c|}{0.601} 
		\\ 
		\multicolumn{1}{|c|}{LRR$_{best}$}       & \multicolumn{1}{c|}{0.747}    & \multicolumn{1}{c|}{0.886}    & \multicolumn{1}{c|}{0.725}   & \multicolumn{1}{c|}{0.789}        & \multicolumn{1}{c|}{0.803}          & \multicolumn{1}{c|}{0.776}       
		& \multicolumn{1}{c|}{0.768}    & \multicolumn{1}{c|}{0.871}    & \multicolumn{1}{c|}{0.736}   & \multicolumn{1}{c|}{0.763}        & \multicolumn{1}{c|}{0.759}          & \multicolumn{1}{c|}{0.767}
		\\ \hline
		\multicolumn{1}{|c|}{Co-reg}        & \multicolumn{1}{c|}{0.771}    & \multicolumn{1}{c|}{0.849}    & \multicolumn{1}{c|}{0.783}   & \multicolumn{1}{c|}{0.829}        & \multicolumn{1}{c|}{0.836}          & \multicolumn{1}{c|}{0.822}       
		& \multicolumn{1}{c|}{0.804}    & \multicolumn{1}{c|}{0.780}    & \multicolumn{1}{c|}{0.755}   & \multicolumn{1}{c|}{0.780}        & \multicolumn{1}{c|}{0.764}          & \multicolumn{1}{c|}{0.798}
		\\ 
		\multicolumn{1}{|c|}{RMSC}        & \multicolumn{1}{c|}{0.808}    & \multicolumn{1}{c|}{0.912}    & \multicolumn{1}{c|}{0.837}   & \multicolumn{1}{c|}{0.871}        & \multicolumn{1}{c|}{0.879}          & \multicolumn{1}{c|}{0.864}       
		& \multicolumn{1}{c|}{0.822}    & \multicolumn{1}{c|}{0.915}    & \multicolumn{1}{c|}{0.789}   & \multicolumn{1}{c|}{0.811}        & \multicolumn{1}{c|}{0.797}          & \multicolumn{1}{c|}{0.826}
		\\ \hline
		\multicolumn{1}{|c|}{DiMSC}        & \multicolumn{1}{c|}{0.814}    & \multicolumn{1}{c|}{0.901}    & \multicolumn{1}{c|}{0.843}   & \multicolumn{1}{c|}{0.880}        & \multicolumn{1}{c|}{0.875}          & \multicolumn{1}{c|}{0.882}       
		& \multicolumn{1}{c|}{0.772}    & \multicolumn{1}{c|}{0.703}    & \multicolumn{1}{c|}{0.652}   & \multicolumn{1}{c|}{0.695}        & \multicolumn{1}{c|}{0.673}          & \multicolumn{1}{c|}{0.718}
		\\ 
		\multicolumn{1}{|c|}{LTMSC}        & \multicolumn{1}{c|}{0.066}    & \multicolumn{1}{c|}{0.379}    & \multicolumn{1}{c|}{0.005}   & \multicolumn{1}{c|}{0.383}        & \multicolumn{1}{c|}{0.239}          & \multicolumn{1}{c|}{0.953}       
		& \multicolumn{1}{c|}{0.775}    & \multicolumn{1}{c|}{0.803}    & \multicolumn{1}{c|}{0.725}   & \multicolumn{1}{c|}{0.753}        & \multicolumn{1}{c|}{0.739}          & \multicolumn{1}{c|}{0.767}
		\\ 
		\multicolumn{1}{|c|}{ECMSC}        & \multicolumn{1}{c|}{0.090}    & \multicolumn{1}{c|}{0.408}    & \multicolumn{1}{c|}{0.060}   & \multicolumn{1}{c|}{0.391}        & \multicolumn{1}{c|}{0.267}          & \multicolumn{1}{c|}{0.942}       
		& \multicolumn{1}{c|}{0.780}    & \multicolumn{1}{c|}{0.718}    & \multicolumn{1}{c|}{0.672}   & \multicolumn{1}{c|}{0.707}        & \multicolumn{1}{c|}{0.660}          & \multicolumn{1}{c|}{0.760}
		\\ 
		\hline
		\multicolumn{1}{|c|}{UR-ETLMSC}        & \multicolumn{1}{c|}{0.808}    & \multicolumn{1}{c|}{0.879}    & \multicolumn{1}{c|}{0.823}   & \multicolumn{1}{c|}{0.865}        & \multicolumn{1}{c|}{0.859}          & \multicolumn{1}{c|}{0.873}       
		& \multicolumn{1}{c|}{0.782}    & \multicolumn{1}{c|}{0.841}    & \multicolumn{1}{c|}{0.719}   & \multicolumn{1}{c|}{0.747}        & \multicolumn{1}{c|}{0.739}          & \multicolumn{1}{c|}{0.756}
		\\ 
		\multicolumn{1}{|c|}{t-SVD-MSC}        & \multicolumn{1}{c|}{0.830}    & \multicolumn{1}{c|}{0.941}    & \multicolumn{1}{c|}{0.853}   & \multicolumn{1}{c|}{0.888}        & \multicolumn{1}{c|}{0.881}          & \multicolumn{1}{c|}{0.896}       
		& \multicolumn{1}{c|}{0.932}    & \multicolumn{1}{c|}{0.955}    & \multicolumn{1}{c|}{0.924}   & \multicolumn{1}{c|}{0.932}        & \multicolumn{1}{c|}{0.930}          & \multicolumn{1}{c|}{0.934}
		\\ 
		\multicolumn{1}{|c|}{ETLMSC}        & \multicolumn{1}{c|}{\textbf{0.984}}    & \multicolumn{1}{c|}{\textbf{0.978}}    & \multicolumn{1}{c|}{\textbf{0.967}}   & \multicolumn{1}{c|}{\textbf{0.977}}        & \multicolumn{1}{c|}{\textbf{0.963}}          & \multicolumn{1}{c|}{\textbf{0.998}}       
		& \multicolumn{1}{c|}{\textbf{0.977}}    & \multicolumn{1}{c|}{\textbf{0.958}}    & \multicolumn{1}{c|}{\textbf{0.953}}   & \multicolumn{1}{c|}{\textbf{0.958}}        & \multicolumn{1}{c|}{\textbf{0.940}}          & \multicolumn{1}{c|}{\textbf{0.980}} 
		\\ \hline
	\end{tabular}
\end{table*}
\begin{table*}[t]
	\centering
	\renewcommand\arraystretch{1.2}
	\caption{Experimental results on the COIL-20 and the Notting-Hill  datasets. For ETLMSC, we set $\lambda=0.003$ and $\lambda=0.0008$ for these two datasets, respectively.}
	\label{tab:coil-notting}
	\begin{tabular}{ccccccccccccc}
		\hline
		\multicolumn{1}{|c|}{Datasets} & \multicolumn{6}{c|}{COIL-20} & \multicolumn{6}{c|}{Notting-Hill}  \\ \hline \hline
		\multicolumn{1}{|c|}{Methods} & \multicolumn{1}{c|}{NMI} & \multicolumn{1}{c|}{ACC} & \multicolumn{1}{c|}{AR} & \multicolumn{1}{c|}{F-score} & \multicolumn{1}{c|}{Precision} & \multicolumn{1}{c|}{Recall} & \multicolumn{1}{c|}{NMI} & \multicolumn{1}{c|}{ACC} & \multicolumn{1}{c|}{AR} & \multicolumn{1}{c|}{F-score} & \multicolumn{1}{c|}{Precision} & \multicolumn{1}{c|}{Recall} \\ \hline
		\hline
		\multicolumn{1}{|c|}{SPC$_{best}$}               
		& \multicolumn{1}{c|}{0.806}    & \multicolumn{1}{c|}{0.672}    & \multicolumn{1}{c|}{0.619}   & \multicolumn{1}{c|}{0.640}        & \multicolumn{1}{c|}{0.596}          & \multicolumn{1}{c|}{0.692} 
		& \multicolumn{1}{c|}{0.723}    & \multicolumn{1}{c|}{0.816}    & \multicolumn{1}{c|}{0.712}   & \multicolumn{1}{c|}{0.775}        & \multicolumn{1}{c|}{0.780}          & \multicolumn{1}{c|}{0.776}
		\\ 
		\multicolumn{1}{|c|}{LRR$_{best}$}              
		& \multicolumn{1}{c|}{0.829}    & \multicolumn{1}{c|}{0.761}    & \multicolumn{1}{c|}{0.720}   & \multicolumn{1}{c|}{0.734}        & \multicolumn{1}{c|}{0.717}          & \multicolumn{1}{c|}{0.751}
		& \multicolumn{1}{c|}{0.579}    & \multicolumn{1}{c|}{0.794}    & \multicolumn{1}{c|}{0.558}   & \multicolumn{1}{c|}{0.653}        & \multicolumn{1}{c|}{0.672}          & \multicolumn{1}{c|}{0.636}
		\\ \hline
		\multicolumn{1}{|c|}{Co-reg}            
		& \multicolumn{1}{c|}{0.774}    & \multicolumn{1}{c|}{0.659}    & \multicolumn{1}{c|}{0.592}   & \multicolumn{1}{c|}{0.613}        & \multicolumn{1}{c|}{0.590}          & \multicolumn{1}{c|}{0.640}
		& \multicolumn{1}{c|}{0.703}    & \multicolumn{1}{c|}{0.805}    & \multicolumn{1}{c|}{0.686}   & \multicolumn{1}{c|}{0.754}        & \multicolumn{1}{c|}{0.766}          & \multicolumn{1}{c|}{0.743}
		\\ 
		\multicolumn{1}{|c|}{RMSC}           
		& \multicolumn{1}{c|}{0.800}    & \multicolumn{1}{c|}{0.685}    & \multicolumn{1}{c|}{0.637}   & \multicolumn{1}{c|}{0.656}        & \multicolumn{1}{c|}{0.620}          & \multicolumn{1}{c|}{0.698}
		& \multicolumn{1}{c|}{0.585}    & \multicolumn{1}{c|}{0.807}    & \multicolumn{1}{c|}{0.496}   & \multicolumn{1}{c|}{0.603}        & \multicolumn{1}{c|}{0.621}          & \multicolumn{1}{c|}{0.586}
		\\ \hline
		\multicolumn{1}{|c|}{DiMSC}        
		& \multicolumn{1}{c|}{0.846}    & \multicolumn{1}{c|}{0.778}    & \multicolumn{1}{c|}{0.732}   & \multicolumn{1}{c|}{0.745}        & \multicolumn{1}{c|}{0.739}          & \multicolumn{1}{c|}{0.751}
		& \multicolumn{1}{c|}{0.799}    & \multicolumn{1}{c|}{0.837}    & \multicolumn{1}{c|}{0.787}   & \multicolumn{1}{c|}{0.834}        & \multicolumn{1}{c|}{0.822}          & \multicolumn{1}{c|}{0.847}
		\\ 
		\multicolumn{1}{|c|}{LTMSC}             
		& \multicolumn{1}{c|}{0.860}    & \multicolumn{1}{c|}{0.804}    & \multicolumn{1}{c|}{0.748}   & \multicolumn{1}{c|}{0.760}        & \multicolumn{1}{c|}{0.741}          & \multicolumn{1}{c|}{0.479}
		& \multicolumn{1}{c|}{0.779}    & \multicolumn{1}{c|}{0.868}    & \multicolumn{1}{c|}{0.777}   & \multicolumn{1}{c|}{0.825}        & \multicolumn{1}{c|}{0.830}          & \multicolumn{1}{c|}{0.814}
		\\ 
		\multicolumn{1}{|c|}{ECMSC}        
		& \multicolumn{1}{c|}{0.942}    & \multicolumn{1}{c|}{0.782}    & \multicolumn{1}{c|}{0.781}   & \multicolumn{1}{c|}{0.794}        & \multicolumn{1}{c|}{0.695}          & \multicolumn{1}{c|}{\textbf{0.925}}
		& \multicolumn{1}{c|}{0.817}    & \multicolumn{1}{c|}{0.767}    & \multicolumn{1}{c|}{0.679}   & \multicolumn{1}{c|}{0.764}        & \multicolumn{1}{c|}{0.637}          & \multicolumn{1}{c|}{0.954}
		\\ \hline
		\multicolumn{1}{|c|}{UR-ETLMSC}        & \multicolumn{1}{c|}{0.829}    & \multicolumn{1}{c|}{0.750}    & \multicolumn{1}{c|}{0.696}   & \multicolumn{1}{c|}{0.711}        & \multicolumn{1}{c|}{0.692}          & \multicolumn{1}{c|}{0.732}       
		& \multicolumn{1}{c|}{0.794}    & \multicolumn{1}{c|}{0.835}    & \multicolumn{1}{c|}{0.787}   & \multicolumn{1}{c|}{0.834}        & \multicolumn{1}{c|}{0.828}          & \multicolumn{1}{c|}{0.840}
		\\ 
		\multicolumn{1}{|c|}{t-SVD-MSC}       
		& \multicolumn{1}{c|}{0.884}    & \multicolumn{1}{c|}{0.830}    & \multicolumn{1}{c|}{0.786}   & \multicolumn{1}{c|}{0.800}        & \multicolumn{1}{c|}{0.785}          & \multicolumn{1}{c|}{0.808}
		& \multicolumn{1}{c|}{{0.900}}    & \multicolumn{1}{c|}{\textbf{0.957}}    & \multicolumn{1}{c|}{\textbf{0.900}}   & \multicolumn{1}{c|}{{0.922}}        & \multicolumn{1}{c|}{{0.937}}          & \multicolumn{1}{c|}{{0.907}}
		\\ 
		\multicolumn{1}{|c|}{ETLMSC}              
		& \multicolumn{1}{c|}{\textbf{0.947}}    & \multicolumn{1}{c|}{\textbf{0.877}}    & \multicolumn{1}{c|}{\textbf{0.862}}   & \multicolumn{1}{c|}{\textbf{0.869}}        & \multicolumn{1}{c|}{\textbf{0.830}}          & \multicolumn{1}{c|}{{0.914}}  
		& \multicolumn{1}{c|}{\textbf{0.911}}    & \multicolumn{1}{c|}{{0.951}}    & \multicolumn{1}{c|}{{0.898}}   & \multicolumn{1}{c|}{\textbf{0.924}}        & \multicolumn{1}{c|}{\textbf{0.940}}          & \multicolumn{1}{c|}{\textbf{0.908}}
		\\ \hline
	\end{tabular}
\end{table*}

\subsubsection{Compared Methods}
We compare our proposed approach ETLMSC and UR-ETLMSC~(the proposed method without tensor rotation) with the following state-of-the-art methods, including two single view and six multi-view methods. 
\\
\textbf{SPC$_{best}$} achieves the best result among all views with standard spectral clustering~\cite{ng2002spectral}.
\\
\textbf{LRR$_{best}$} achieves the best result among all views with the low-rank representation~\cite{liu2013robust}.
\\
\textbf{Co-reg}~\cite{kumar2011coreg} is the co-regularization method for spectral clustering, which co-regularizes the clustering hypothesis to explore the complementary information.
\\
\textbf{RMSC}~\cite{RMSC} recovers a shared low-rank transition probability matrix as input to the Markov chain based spectral clustering.
\\
\textbf{DiMSC}~\cite{cao2015diversity} employs the HSIC as a diversity term to explore the complementarity of multi-view representations. 
\\
\textbf{LTMSC}~\cite{zhang2015low} adopts the low-rank tensor constraint for multi-view subspace clustering.
\\
\textbf{ECMSC}~\cite{wang2017exclusivity} consists of position-aware exclusivity term and consistency term for regularization.
\\
\textbf{t-SVD-MSC}~\cite{xie2016unifying} uses the t-SVD based tensor nuclear norm to learn optimal subspace.
\par
Among all above methods, only SPC$_{best}$, Co-reg, and RMSC are spectral clustering methods, and other methods are self-representation based subspace clustering methods.

\subsubsection{Evaluation Metrics}
To comprehensively evaluate the performance of clustering, we adopt all six commonly used metrics including normalized mutual information~(NMI), accuracy~(ACC), adjusted rand index~(AR), F-score~, precision and recall.
These six metrics favour different properties in clustering task. 
For all metrics, the higher value indicates the better performance.
\subsection{Experimental Results and Analysis}
\begin{table*}[t]
	\centering
	\renewcommand\arraystretch{1.2}
	\caption{Experimental results on the Scene-15 and the MITIndoor-67 datasets. For ETLMSC, we set $\lambda=0.003$ for both two datasets.}
	\label{tab:scene-mit}
	\begin{tabular}{ccccccccccccc}
		\hline
		\multicolumn{1}{|c|}{Datasets} & \multicolumn{6}{c|}{Scene-15} & \multicolumn{6}{c|}{MITIndoor-67}  \\ \hline \hline
		\multicolumn{1}{|c|}{Methods} & \multicolumn{1}{c|}{NMI} & \multicolumn{1}{c|}{ACC} & \multicolumn{1}{c|}{AR} & \multicolumn{1}{c|}{F-score} & \multicolumn{1}{c|}{Precision} & \multicolumn{1}{c|}{Recall} & \multicolumn{1}{c|}{NMI} & \multicolumn{1}{c|}{ACC} & \multicolumn{1}{c|}{AR} & \multicolumn{1}{c|}{F-score} & \multicolumn{1}{c|}{Precision} & \multicolumn{1}{c|}{Recall} \\ \hline
		\multicolumn{1}{|c|}{SPC$_{best}$}        & \multicolumn{1}{c|}{0.421}    & \multicolumn{1}{c|}{0.437}    & \multicolumn{1}{c|}{0.270}   & \multicolumn{1}{c|}{0.321}        & \multicolumn{1}{c|}{0.314}          & \multicolumn{1}{c|}{0.329} 
		& \multicolumn{1}{c|}{0.559}    & \multicolumn{1}{c|}{0.443}    & \multicolumn{1}{c|}{0.304}   & \multicolumn{1}{c|}{0.315}        & \multicolumn{1}{c|}{0.294}          & \multicolumn{1}{c|}{0.340}      \\ 
		\multicolumn{1}{|c|}{LRR$_{best}$}       & \multicolumn{1}{c|}{0.426}    & \multicolumn{1}{c|}{0.445}    & \multicolumn{1}{c|}{0.272}   & \multicolumn{1}{c|}{0.324}        & \multicolumn{1}{c|}{0.316}          & \multicolumn{1}{c|}{0.333}       
		& \multicolumn{1}{c|}{0.226}    & \multicolumn{1}{c|}{0.120}    & \multicolumn{1}{c|}{0.031}   & \multicolumn{1}{c|}{0.045}        & \multicolumn{1}{c|}{0.044}          & \multicolumn{1}{c|}{0.047}\\ \hline
		\multicolumn{1}{|c|}{Co-reg}        & \multicolumn{1}{c|}{0.470}    & \multicolumn{1}{c|}{0.503}    & \multicolumn{1}{c|}{0.334}   & \multicolumn{1}{c|}{0.380}        & \multicolumn{1}{c|}{0.382}          & \multicolumn{1}{c|}{0.378}       
		& \multicolumn{1}{c|}{0.270}    & \multicolumn{1}{c|}{0.149}    & \multicolumn{1}{c|}{0.054}   & \multicolumn{1}{c|}{0.067}        & \multicolumn{1}{c|}{0.066}          & \multicolumn{1}{c|}{0.070}\\ 
		\multicolumn{1}{|c|}{RMSC}        & \multicolumn{1}{c|}{0.564}    & \multicolumn{1}{c|}{0.507}    & \multicolumn{1}{c|}{0.394}   & \multicolumn{1}{c|}{0.437}        & \multicolumn{1}{c|}{0.425}          & \multicolumn{1}{c|}{0.450}      
		& \multicolumn{1}{c|}{0.342}    & \multicolumn{1}{c|}{0.232}    & \multicolumn{1}{c|}{0.110}   & \multicolumn{1}{c|}{0.123}        & \multicolumn{1}{c|}{0.121}          & \multicolumn{1}{c|}{0.125}
		\\ \hline
		\multicolumn{1}{|c|}{DiMSC}        & \multicolumn{1}{c|}{0.269}    & \multicolumn{1}{c|}{0.300}    & \multicolumn{1}{c|}{0.117}   & \multicolumn{1}{c|}{0.181}        & \multicolumn{1}{c|}{0.173}          & \multicolumn{1}{c|}{0.190}       
		& \multicolumn{1}{c|}{0.383}    & \multicolumn{1}{c|}{0.246}    & \multicolumn{1}{c|}{0.128}   & \multicolumn{1}{c|}{0.141}        & \multicolumn{1}{c|}{0.138}          & \multicolumn{1}{c|}{0.144}\\ 
		\multicolumn{1}{|c|}{LTMSC}       & \multicolumn{1}{c|}{0.571}    & \multicolumn{1}{c|}{0.574}    & \multicolumn{1}{c|}{0.424}   & \multicolumn{1}{c|}{0.465}        & \multicolumn{1}{c|}{0.452}          & \multicolumn{1}{c|}{0.479}       
		& \multicolumn{1}{c|}{0.226}    & \multicolumn{1}{c|}{0.120}    & \multicolumn{1}{c|}{0.031}   & \multicolumn{1}{c|}{0.045}        & \multicolumn{1}{c|}{0.044}          & \multicolumn{1}{c|}{0.047}
		\\ 
		\multicolumn{1}{|c|}{ECMSC}       & \multicolumn{1}{c|}{0.463}    & \multicolumn{1}{c|}{0.457}    & \multicolumn{1}{c|}{0.303}   & \multicolumn{1}{c|}{0.357}        & \multicolumn{1}{c|}{0.318}          & \multicolumn{1}{c|}{0.408}      
		& \multicolumn{1}{c|}{0.590}    & \multicolumn{1}{c|}{0.469}    & \multicolumn{1}{c|}{0.323}   & \multicolumn{1}{c|}{0.333}        & \multicolumn{1}{c|}{0.314}          & \multicolumn{1}{c|}{0.355}
		\\ \hline
		\multicolumn{1}{|c|}{UR-ETLMSC}        & \multicolumn{1}{c|}{0.536}    & \multicolumn{1}{c|}{0.534}    & \multicolumn{1}{c|}{0.369}   & \multicolumn{1}{c|}{0.419}        & \multicolumn{1}{c|}{0.420}          & \multicolumn{1}{c|}{0.419}       
		& \multicolumn{1}{c|}{0.467}    & \multicolumn{1}{c|}{0.335}    & \multicolumn{1}{c|}{0.204}   & \multicolumn{1}{c|}{0.216}        & \multicolumn{1}{c|}{0.211}          & \multicolumn{1}{c|}{0.220}
		\\
		\multicolumn{1}{|c|}{t-SVD-MSC}        & \multicolumn{1}{c|}{0.858}    & \multicolumn{1}{c|}{0.812}    & \multicolumn{1}{c|}{0.771}   & \multicolumn{1}{c|}{0.788}        & \multicolumn{1}{c|}{0.743}          & \multicolumn{1}{c|}{0.839}     
		& \multicolumn{1}{c|}{0.750}    & \multicolumn{1}{c|}{0.684}    & \multicolumn{1}{c|}{0.555}   & \multicolumn{1}{c|}{0.562}        & \multicolumn{1}{c|}{0.543}          & \multicolumn{1}{c|}{0.582}
		\\ 
		\multicolumn{1}{|c|}{ETLMSC}        & \multicolumn{1}{c|}{\textbf{0.902}}    & \multicolumn{1}{c|}{\textbf{0.878}}    & \multicolumn{1}{c|}{\textbf{0.851}}   & \multicolumn{1}{c|}{\textbf{0.862}}        & \multicolumn{1}{c|}{\textbf{0.848}}          & \multicolumn{1}{c|}{\textbf{0.877}}       
		& \multicolumn{1}{c|}{\textbf{0.899}}    & \multicolumn{1}{c|}{\textbf{0.775}}    & \multicolumn{1}{c|}{\textbf{0.729}}   & \multicolumn{1}{c|}{\textbf{0.733}}        & \multicolumn{1}{c|}{\textbf{0.709}}          & \multicolumn{1}{c|}{\textbf{0.758}}
		\\ \hline
	\end{tabular}
\end{table*}

\begin{table}[t]
	\centering
	\renewcommand\arraystretch{1.2}
	\caption{Experimental results on the Caltech-101 datasets. For ETLMSC, we set $\lambda=0.003$.}
	\label{tab:caltech}
	\begin{tabular}{ccccccc}
		\hline
		\multicolumn{1}{|c|}{Datasets} & \multicolumn{6}{c|}{Caltech-101}  \\ \hline \hline
		\multicolumn{1}{|c|}{Methods}  & \multicolumn{1}{c|}{NMI} & \multicolumn{1}{c|}{ACC} & \multicolumn{1}{c|}{AR} & \multicolumn{1}{c|}{F-score} & \multicolumn{1}{c|}{Precision} & \multicolumn{1}{c|}{Recall} \\ \hline
		\multicolumn{1}{|c|}{SPC$_{best}$}           
		& \multicolumn{1}{c|}{0.723}    & \multicolumn{1}{c|}{0.484}    & \multicolumn{1}{c|}{0.319}   & \multicolumn{1}{c|}{0.340}        & \multicolumn{1}{c|}{0.597}          & \multicolumn{1}{c|}{0.235} 
		\\ 
		\multicolumn{1}{|c|}{LRR$_{best}$}       
		& \multicolumn{1}{c|}{0.728}    & \multicolumn{1}{c|}{0.510}    & \multicolumn{1}{c|}{0.304}   & \multicolumn{1}{c|}{0.339}        & \multicolumn{1}{c|}{0.627}          & \multicolumn{1}{c|}{0.231}       
		\\ 
		\hline
		\multicolumn{1}{|c|}{Co-reg}        
		& \multicolumn{1}{c|}{0.824}    & \multicolumn{1}{c|}{0.582}    & \multicolumn{1}{c|}{0.401}   & \multicolumn{1}{c|}{0.412}        & \multicolumn{1}{c|}{0.661}          & \multicolumn{1}{c|}{0.301}       
		\\ 
		\multicolumn{1}{|c|}{RMSC}        
		& \multicolumn{1}{c|}{0.573}    & \multicolumn{1}{c|}{0.346}    & \multicolumn{1}{c|}{0.246}   & \multicolumn{1}{c|}{0.258}        & \multicolumn{1}{c|}{0.457}          & \multicolumn{1}{c|}{0.182}      
		\\ \hline
		\multicolumn{1}{|c|}{DiMSC}        
		& \multicolumn{1}{c|}{0.589}    & \multicolumn{1}{c|}{0.351}    & \multicolumn{1}{c|}{0.226}   & \multicolumn{1}{c|}{0.253}        & \multicolumn{1}{c|}{0.362}          & \multicolumn{1}{c|}{0.191}       
		\\ 
		\multicolumn{1}{|c|}{LTMSC}       
		& \multicolumn{1}{c|}{0.788}    & \multicolumn{1}{c|}{0.559}    & \multicolumn{1}{c|}{0.393}   & \multicolumn{1}{c|}{0.403}        & \multicolumn{1}{c|}{0.670}          & \multicolumn{1}{c|}{0.288}       
		\\ 
		\multicolumn{1}{|c|}{ECMSC}       
		& \multicolumn{1}{c|}{0.662}    & \multicolumn{1}{c|}{0.419}    & \multicolumn{1}{c|}{0.312}   & \multicolumn{1}{c|}{0.326}        & \multicolumn{1}{c|}{0.465}          & \multicolumn{1}{c|}{0.251}      
		\\ \hline
		\multicolumn{1}{|c|}{UR-ETLMSC}        
		& \multicolumn{1}{c|}{0.740}    & \multicolumn{1}{c|}{0.463}    & \multicolumn{1}{c|}{0.342}   & \multicolumn{1}{c|}{0.352}        & \multicolumn{1}{c|}{0.638}          & \multicolumn{1}{c|}{0.243}     
		\\ 
		\multicolumn{1}{|c|}{t-SVD-MSC}        
		& \multicolumn{1}{c|}{0.858}    & \multicolumn{1}{c|}{0.607}    & \multicolumn{1}{c|}{0.430}   & \multicolumn{1}{c|}{0.440}        & \multicolumn{1}{c|}{0.742}          & \multicolumn{1}{c|}{0.323}     
		\\ 
		\multicolumn{1}{|c|}{ETLMSC}        
		& \multicolumn{1}{c|}{\textbf{0.899}}    & \multicolumn{1}{c|}{\textbf{0.639}}    & \multicolumn{1}{c|}{\textbf{0.456}}   & \multicolumn{1}{c|}{\textbf{0.465}}        & \multicolumn{1}{c|}{\textbf{0.825}}          & \multicolumn{1}{c|}{\textbf{0.324}}       
		\\ \hline
	\end{tabular}
\end{table}
\begin{figure*}[t]
	\centering 
	\subfigure[LTMSC]{ 
		{\label{fig:SCENE15_LTMSC}} 
		\includegraphics[width=2.42in]{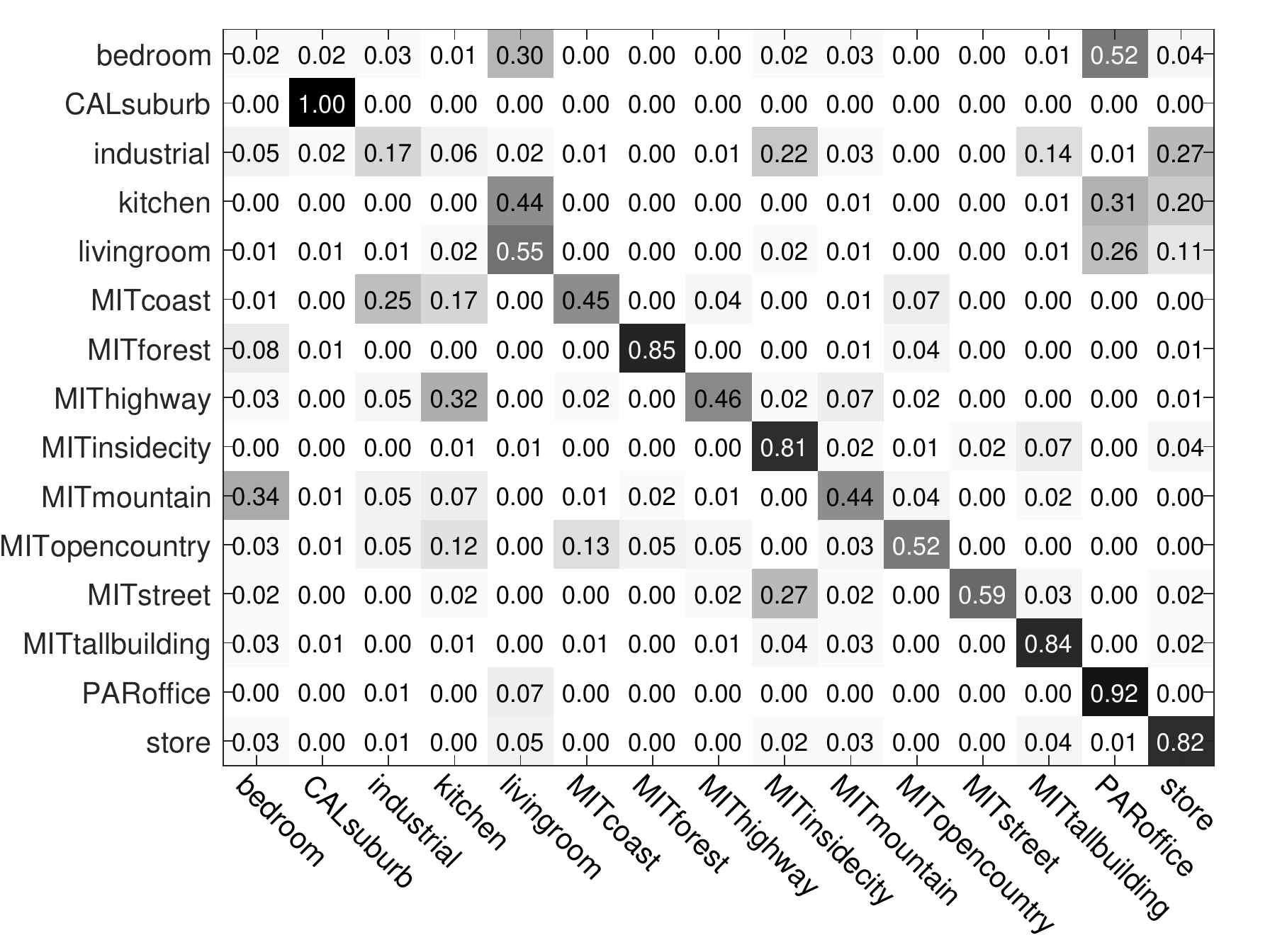}} 
	\hspace{-8mm} 
	\subfigure[t-SVD-MSC]{ 
		{\label{fig:SCENE15_TSVDMSC}} 
		\includegraphics[width=2.33in]{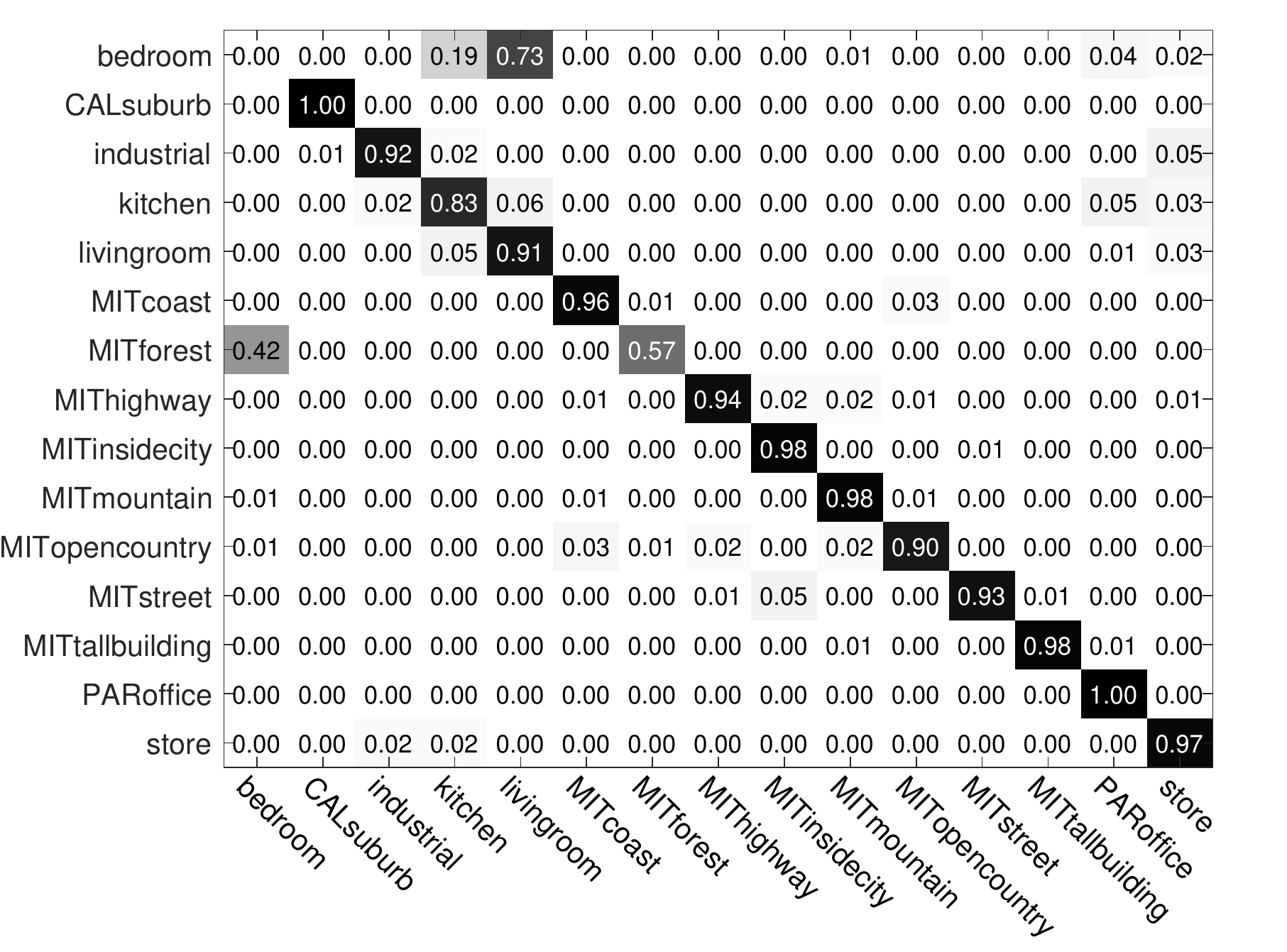}} 
	\hspace{-4mm} 
	\subfigure[ETLMSC]{ 
		\label{fig:SCENE15_ETLMSC} 
		\includegraphics[width=2.25in]{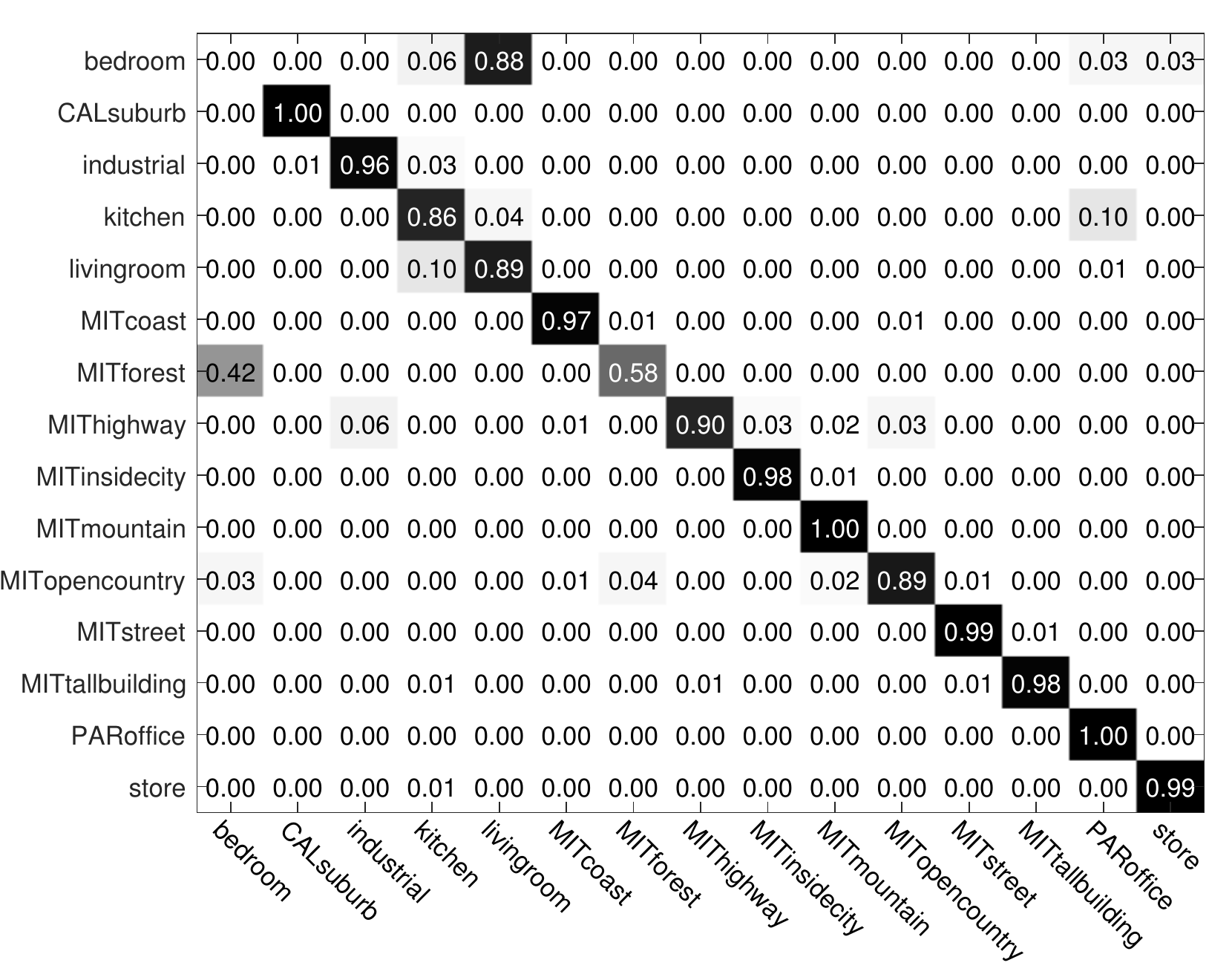}} 
	\caption{The confusion matrices comparison among three tensor based methods, including the proposed LTMSC, t-SVD-MSC, and ETLMSC on the  Scene-15 dataset.} 
	\label{fig:conf_matrix_scene15} 
\end{figure*}
\subsubsection{Performance Comparison}
We present the detailed clustering results on seven datasets in Tables~\ref{tab:bbc-uci}-\ref{tab:caltech}.
All results are measured by the average of $20$ runs.
In each table, the bold values represent the best performance.
To better compare the performance of different methods, we divide all methods into four subclasses in the table, including single view methods, spectral clustering methods, subspace learning methods, and tensor based methods.
The optimal parameters for these methods are fine-tuned by grid searching.
\par
On all datasets, t-SVD-MSC and the proposed ETLMSC achieve the top two best results under nearly all these different metrics.
From Tables~\ref{tab:bbc-uci}-\ref{tab:caltech}, 
we can easily see that our proposed ETLMSC achieves the best performance on the BBC-Sport, UCI-Digits, COIL-20, Scene-15, MITIndoor-67, and Caltech-101 datasets under all six evaluation metrics. 
Especially on the BBC-Sport and MITIndoor-67 datasets, our results are more than $10\%$ higher than the second best results achieved by t-SVD-MSC.
There are also $2\%$, $2\%$, $6\%$ and $3\%$ improvement compared with the second best performance of t-SVD-MSC on the UCI-Digits, COIL-20, Scene-15, and Caltech-101 datasets, respectively.
The Notting-Hill dataset is a video based face dataset.
According to~\cite{arpit2014dimensionality,zhang2014jointly}, facial images have the subspace structure,
and self-representation based subspace learning method is more suitable for this task.
While t-SVD-MSC is based on subspace learning, 
the performance of our method is still comparable to that achieved by t-SVD-MSC, and much higher than those of all other methods, which is shown in the right part of Table~\ref{tab:coil-notting}.
\par
For single view methods, they obtain good performance.
But in general, multi-view methods work better than single view methods.
Moreover, both ECMSC and DiMSC work very well for this task. As they both try to investigate complementary information, it shows that it is necessary to learn view-specific information.
\par
Tensor based methods, including ETLMSC and t-SVD-MSC, achieve significant improvement compared with all other state-of-the-art methods in most cases. 
There is a huge gap between tensor based methods and other methods, which can be attributed to the effectiveness of tensor based correlations exploration.
In Fig.~\ref{fig:conf_matrix_scene15}, we also present the confusion matrices of these three tensor based methods on the Scene-15 dataset. The row and column names correspond to the ground-truth and predicted labels, respectively. We can see that compared with LTMSC, our proposed ETLMSC and t-SVD-MSC achieve much better results in almost all classes in terms of accuracy, which can be attribute to the effectiveness of t-SVD decomposition based tensor nuclear norm.  
Compared with t-SVD-MSC, our ETLMSC improves slightly in many categories, which can also be verified by the accuracy.
\par
\begin{figure*}[htp]
	\centering 
	\subfigure[BBC-Sport]{ 
		{\label{fig:lambda_BBC}} 
		\includegraphics[width=2.2in]{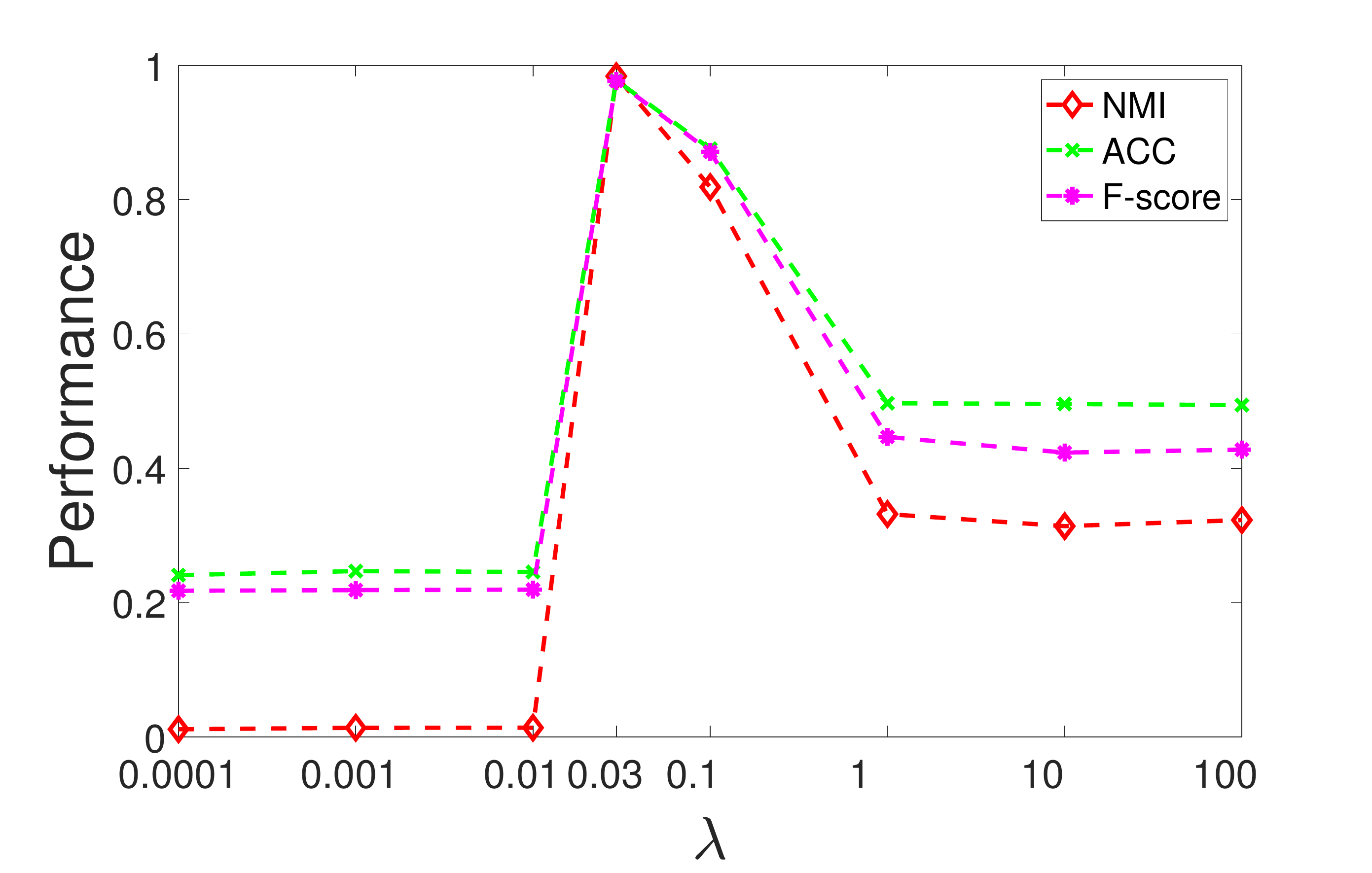}} 
	\hspace{-2.6mm} 
	\subfigure[UCI-Digits]{ 
		\label{fig:lambda_Digit} 
		\includegraphics[width=2.2in]{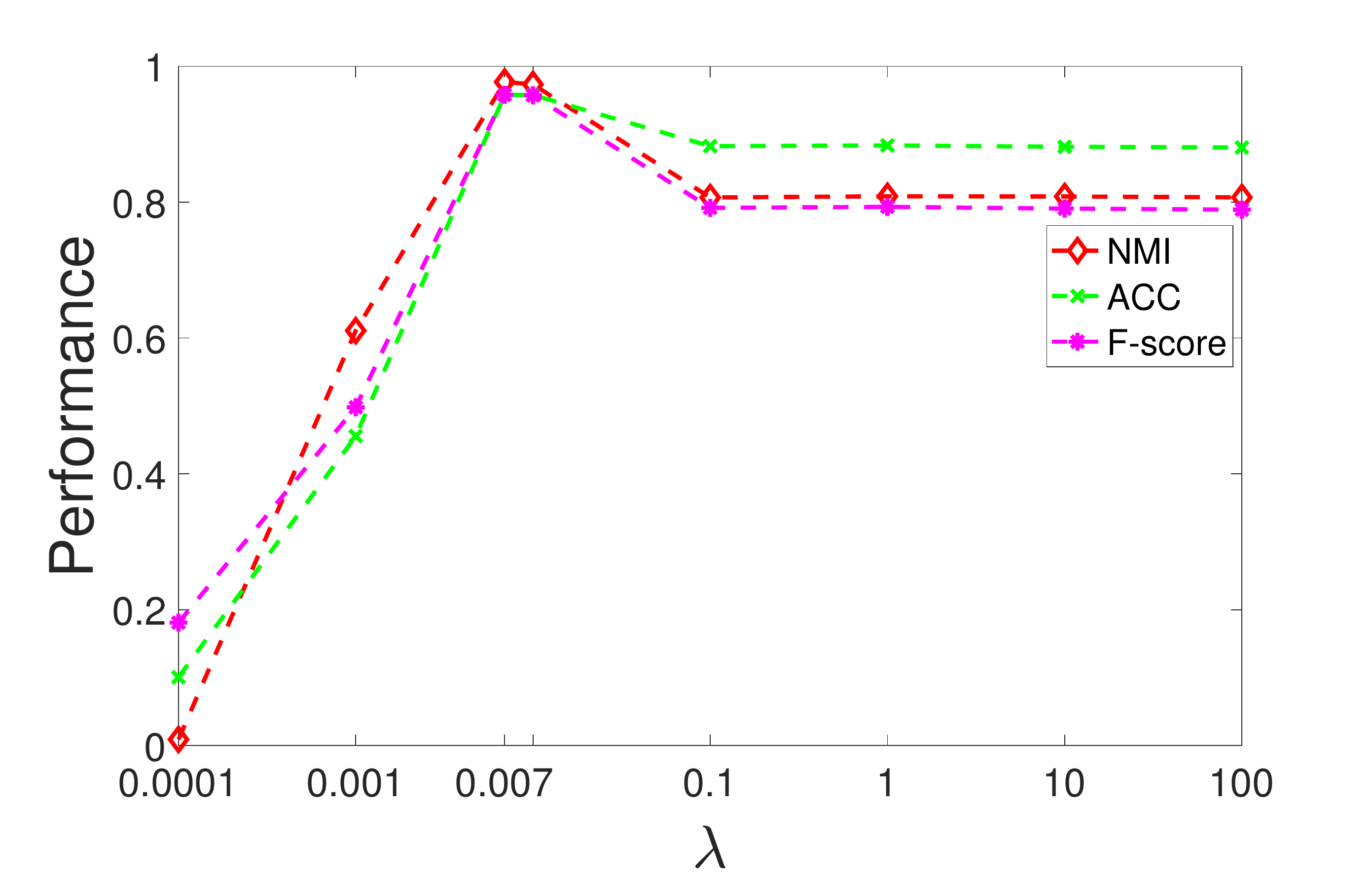}} 
	\hspace{-2.5mm} 
	\subfigure[COIL-20]{ 
		\label{fig:lambda_COIL20} 
		\includegraphics[width=2.3in]{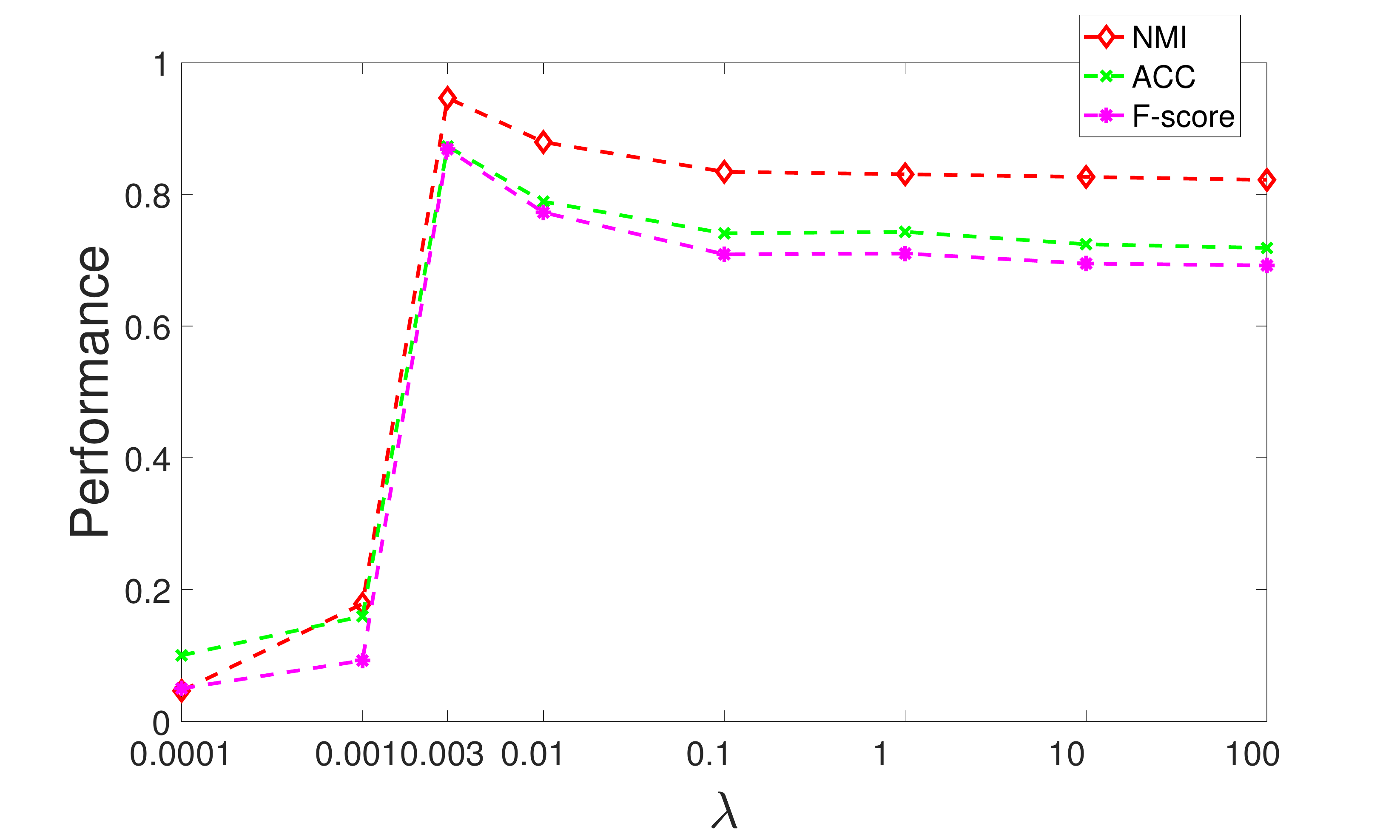}}
	\hspace{-2mm} 
	\subfigure[Notting-Hill]{ 
		\label{fig:lambda_NH} 
		\includegraphics[width=2.23in]{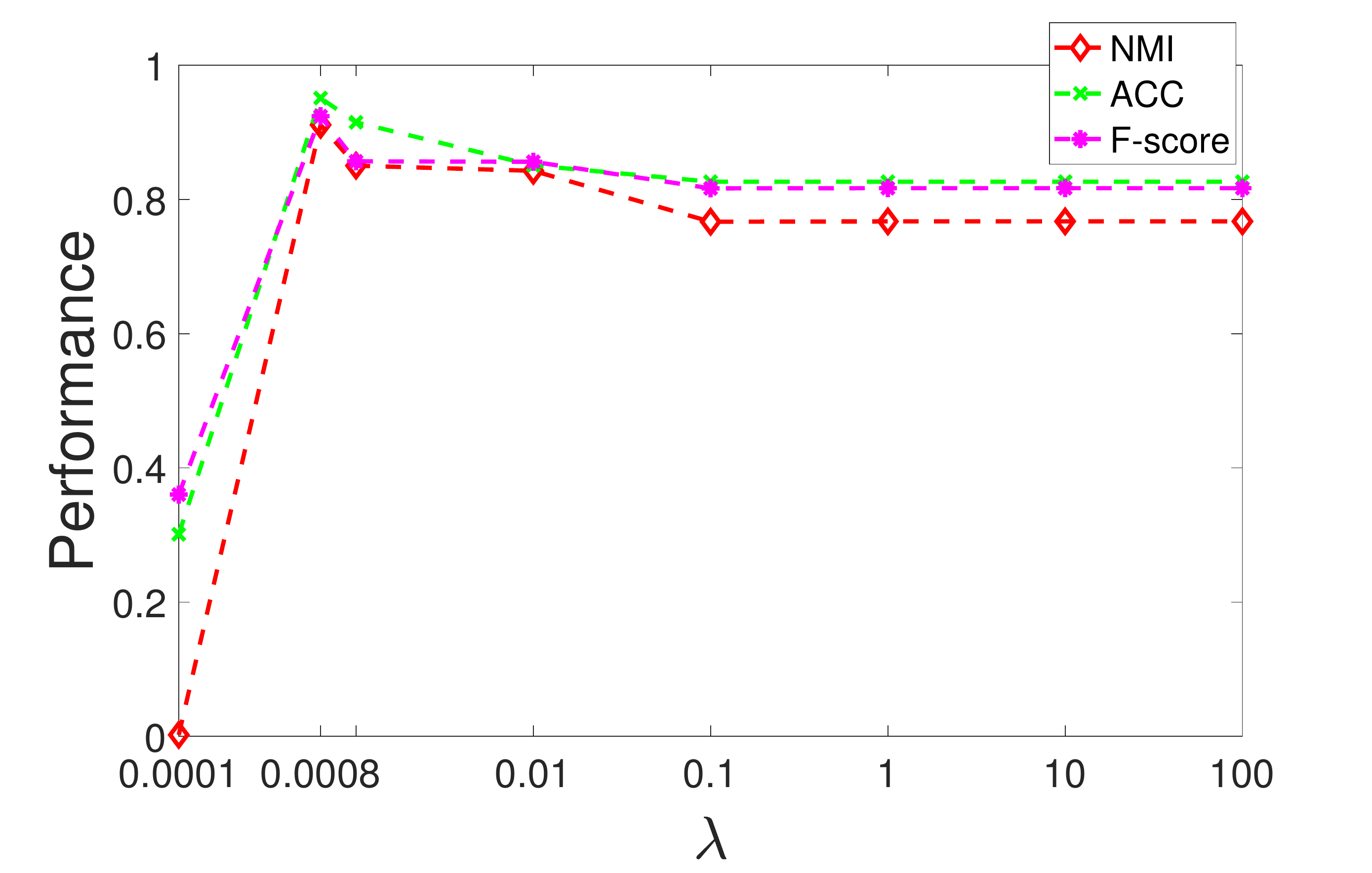}}
	\hspace{-2mm} 
	\subfigure[Scene-15]{ 
		\label{fig:lambda_Scene} 
		\includegraphics[width=2.2in]{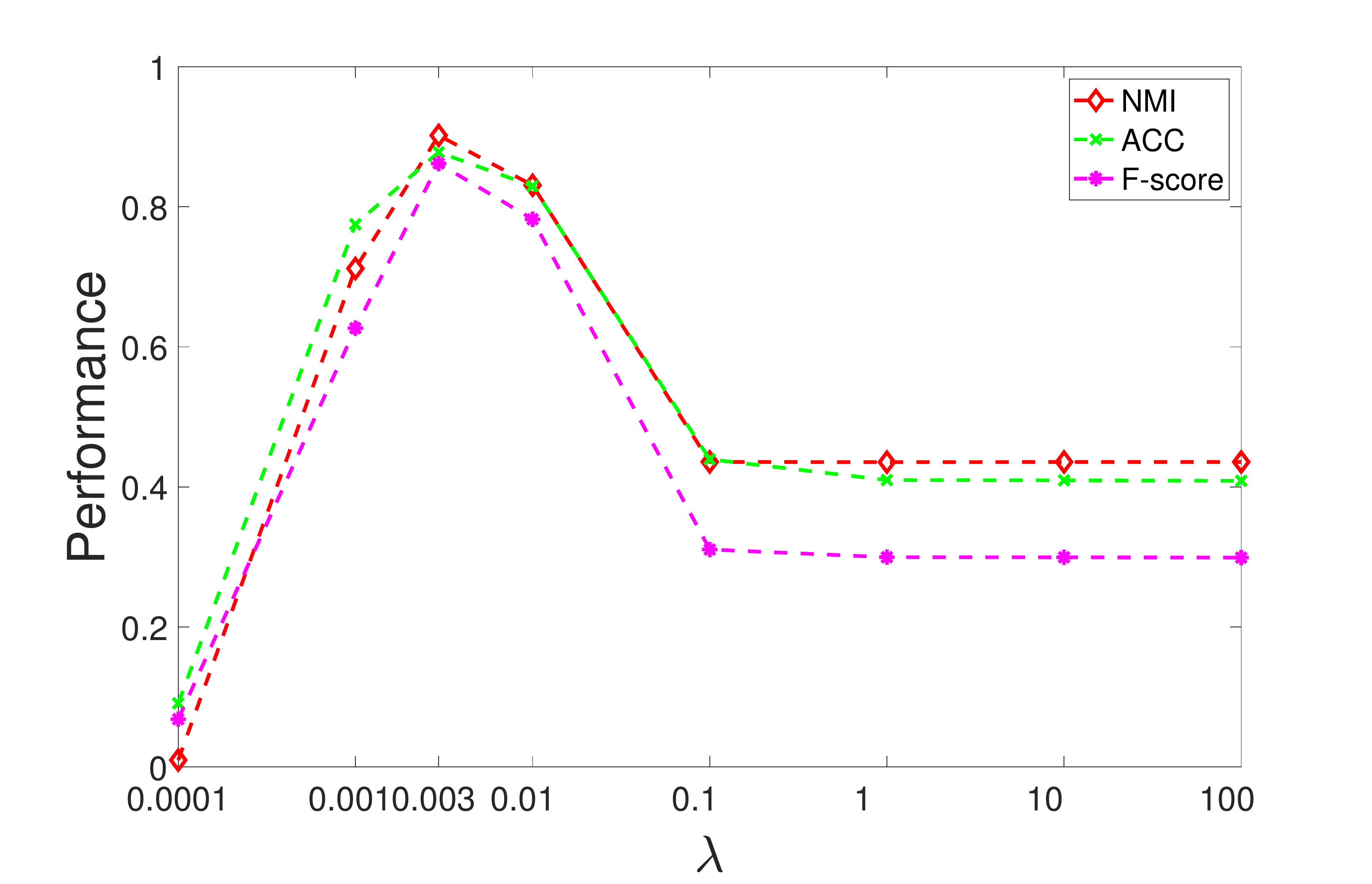}}
	\hspace{-2mm} 
	\subfigure[MITIndoor-67]{ 
		\label{fig:lambda_MIT} 
		\includegraphics[width=2.26in]{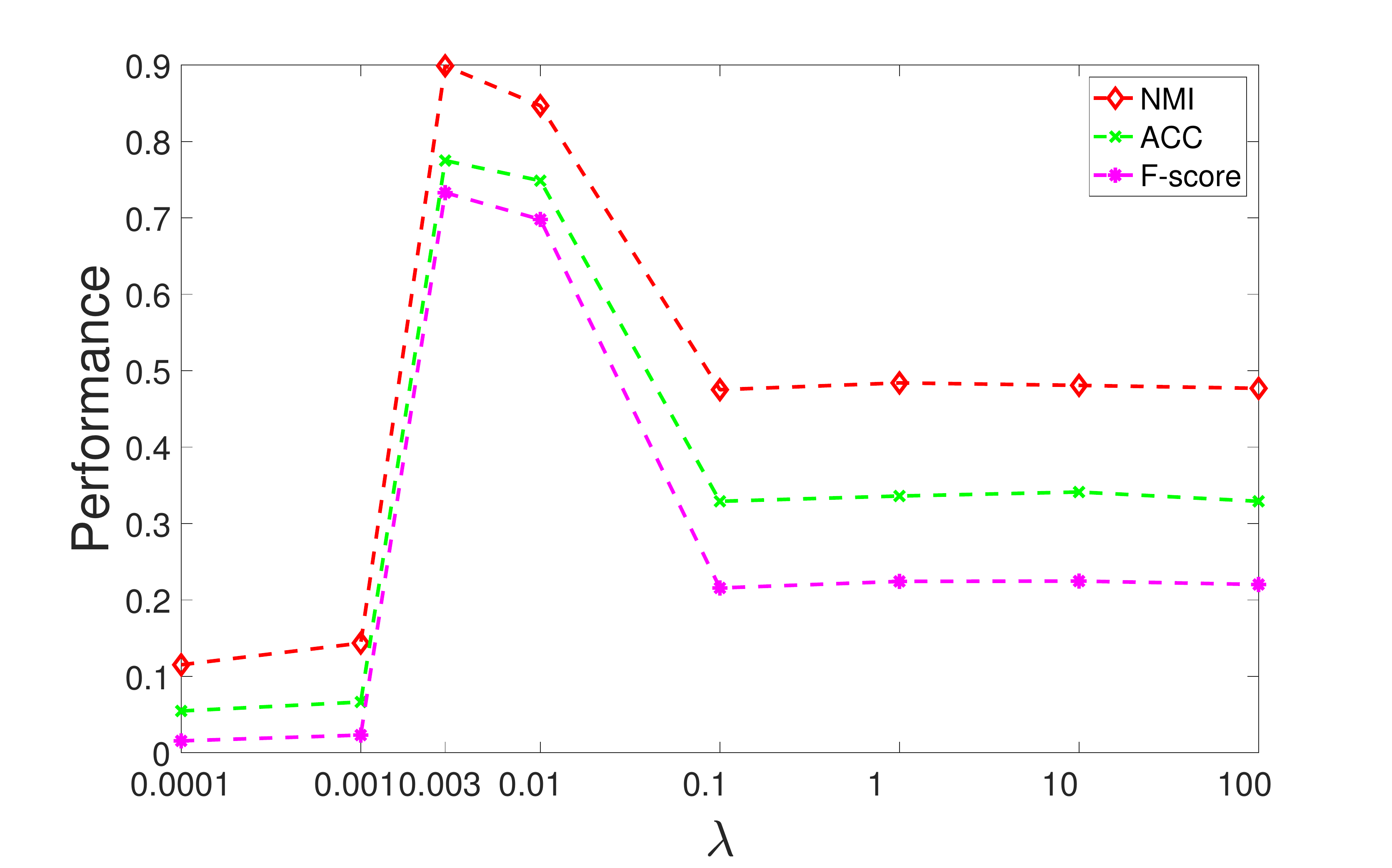}}
	\caption{Parameter tuning with respect to $\lambda$ on the first six datasets. Please note that the x-axis is in log scale.} 
	\label{fig:lambda} 
\end{figure*}
\begin{figure*}[htp]
	\centering 
	\subfigure[BBC-Sport]{ 
		{\label{fig:sigma_BBC}} 
		\includegraphics[width=2.2in]{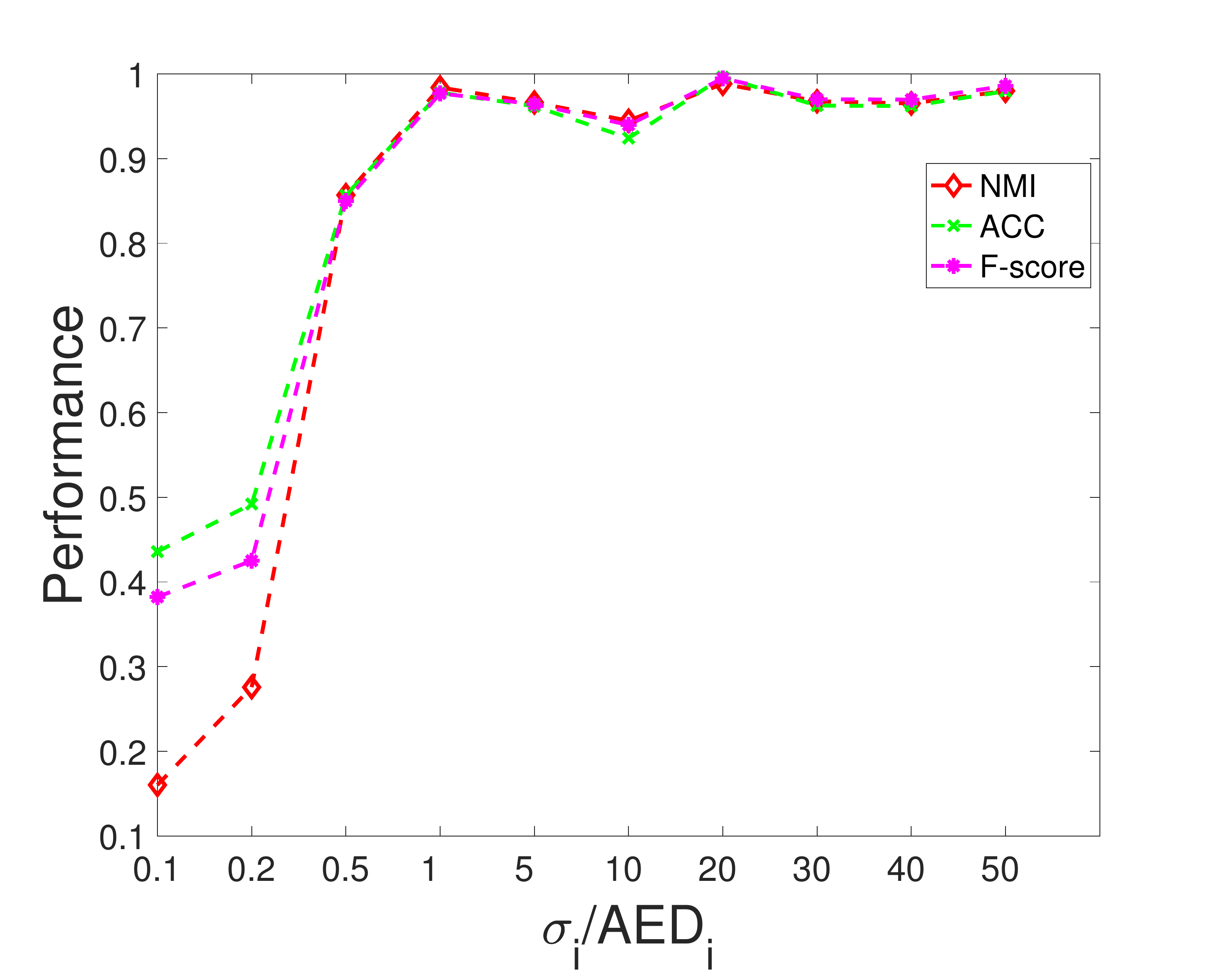}} 
	\hspace{-2.6mm} 
	\subfigure[UCI-Digits]{ 
		\label{fig:sigma_Digit} 
		\includegraphics[width=2.2in]{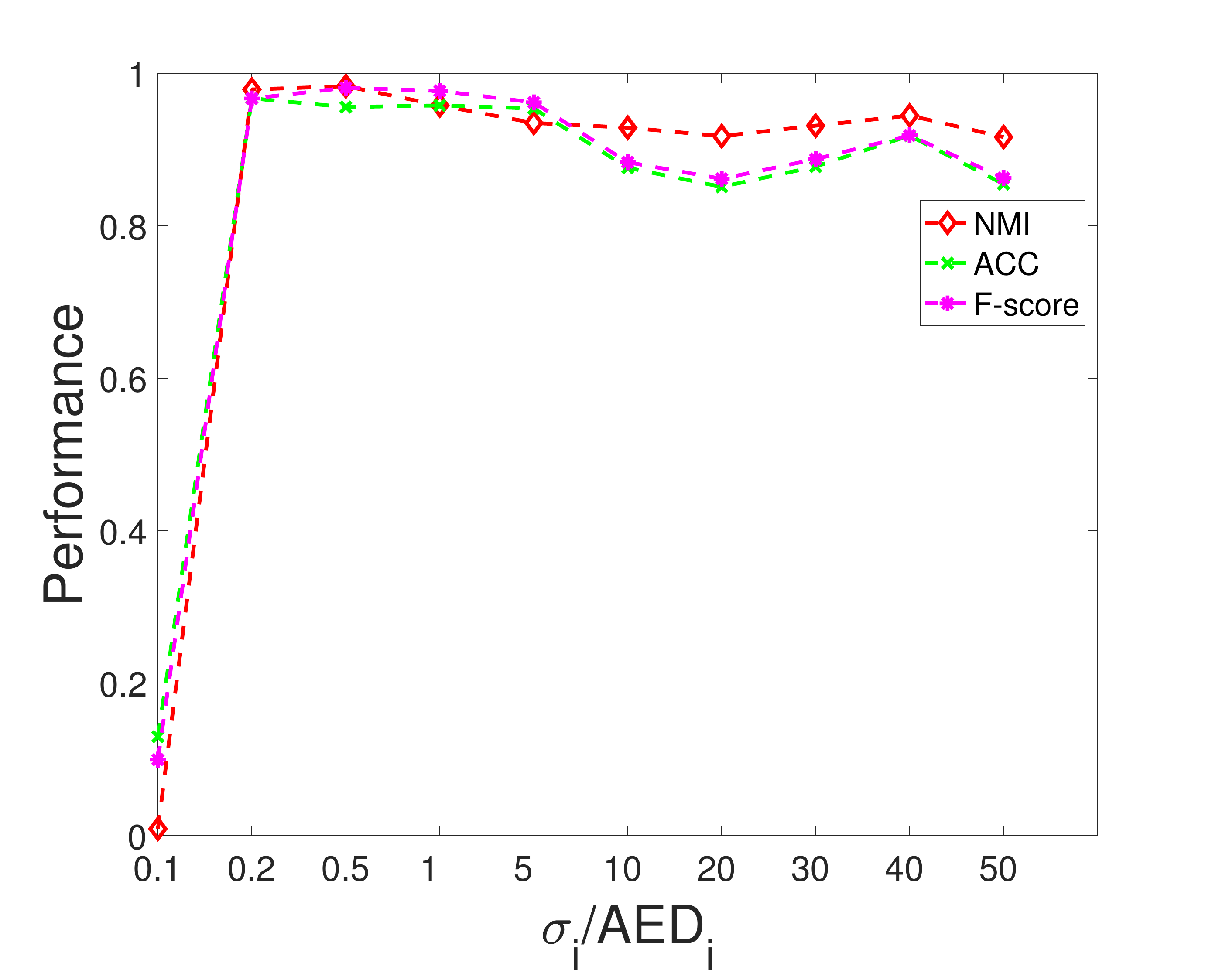}} 
	\hspace{-2.5mm} 
	\subfigure[COIL-20]{ 
		\label{fig:sigma_COIL20} 
		\includegraphics[width=2.2in]{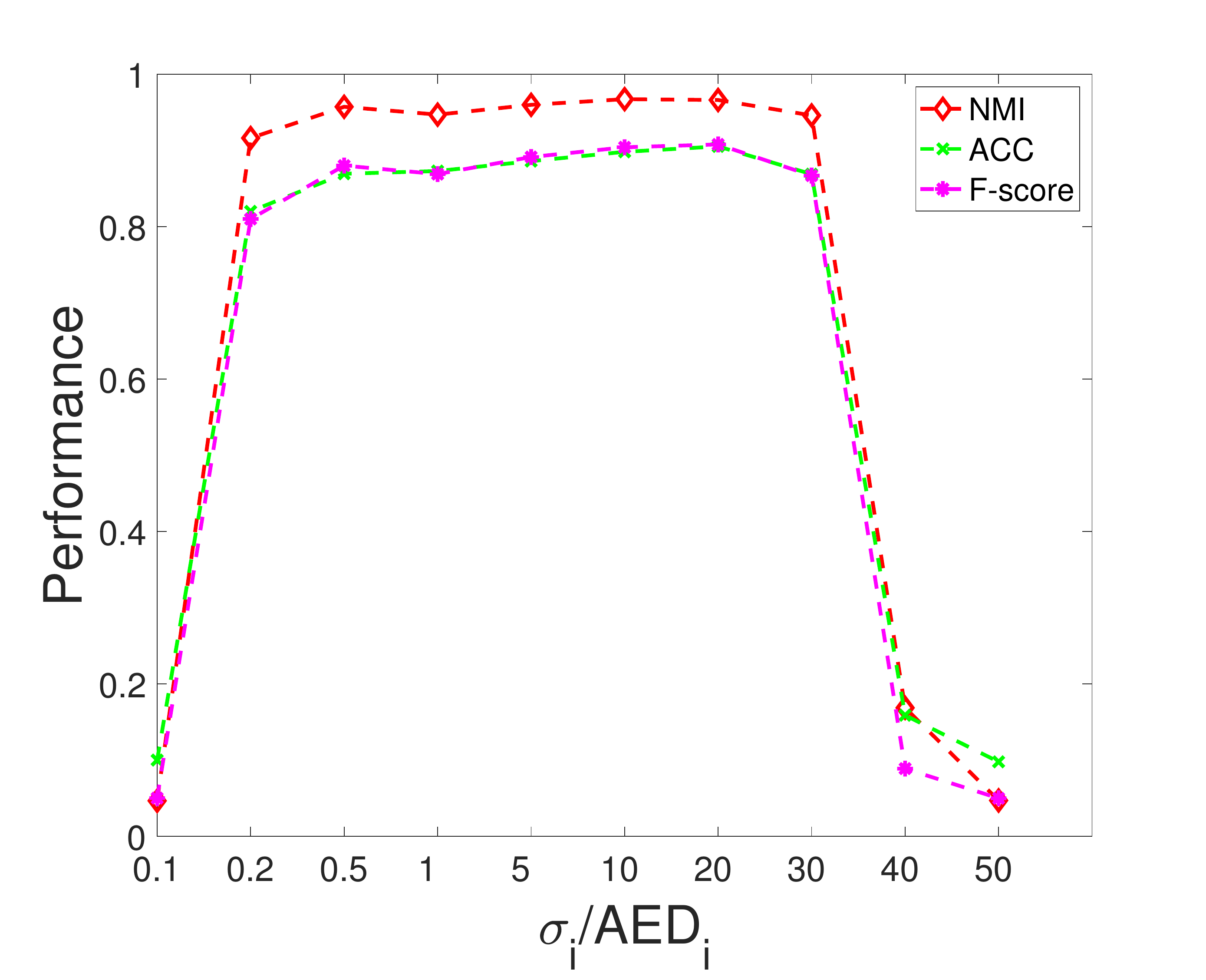}}
	\hspace{-2mm} 
	\subfigure[Notting-Hill]{ 
		\label{fig:sigma_NH} 
		\includegraphics[width=2.2in]{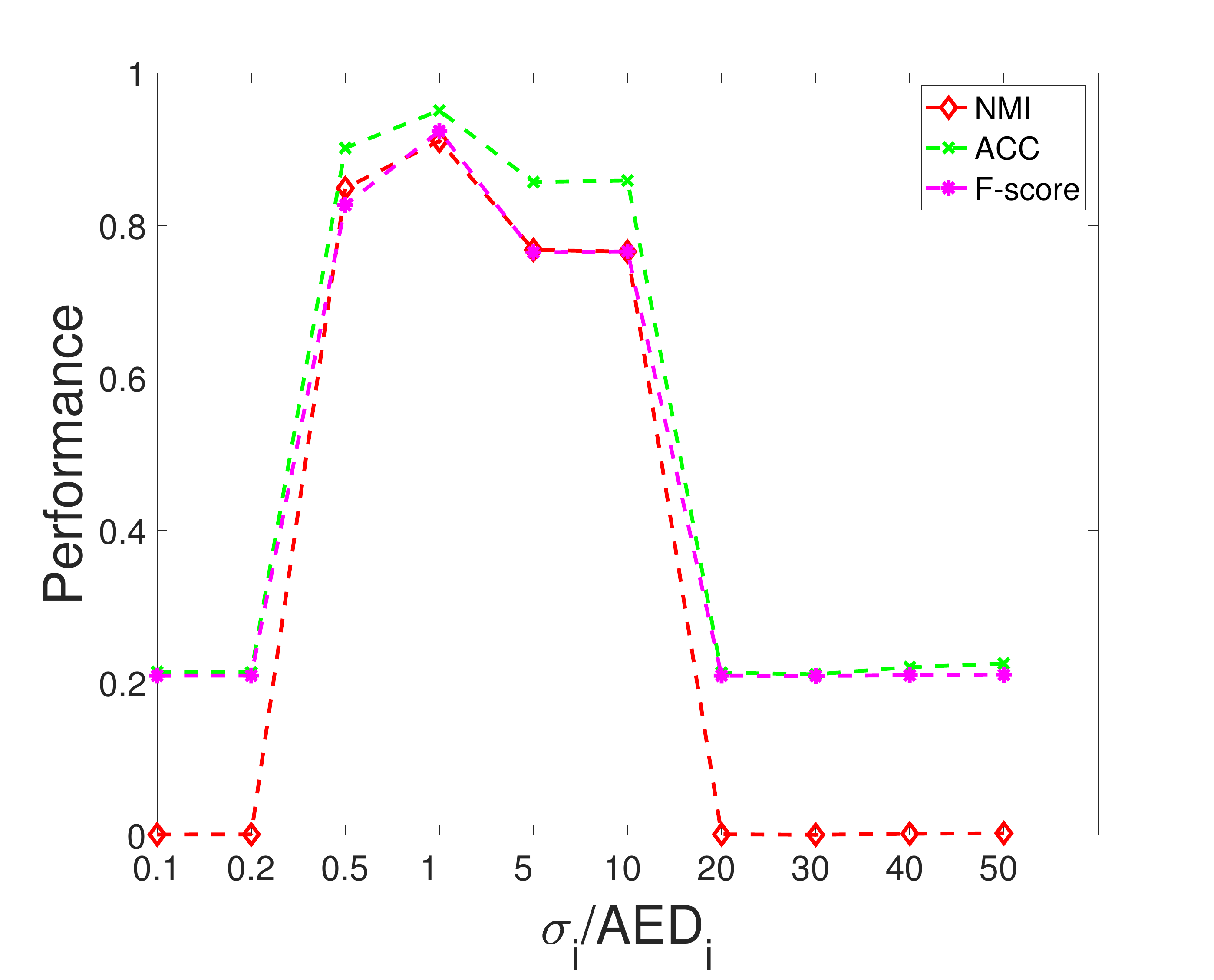}}
	\hspace{-2mm} 
	\subfigure[Scene-15]{ 
		\label{fig:sigma_Scene} 
		\includegraphics[width=2.2in]{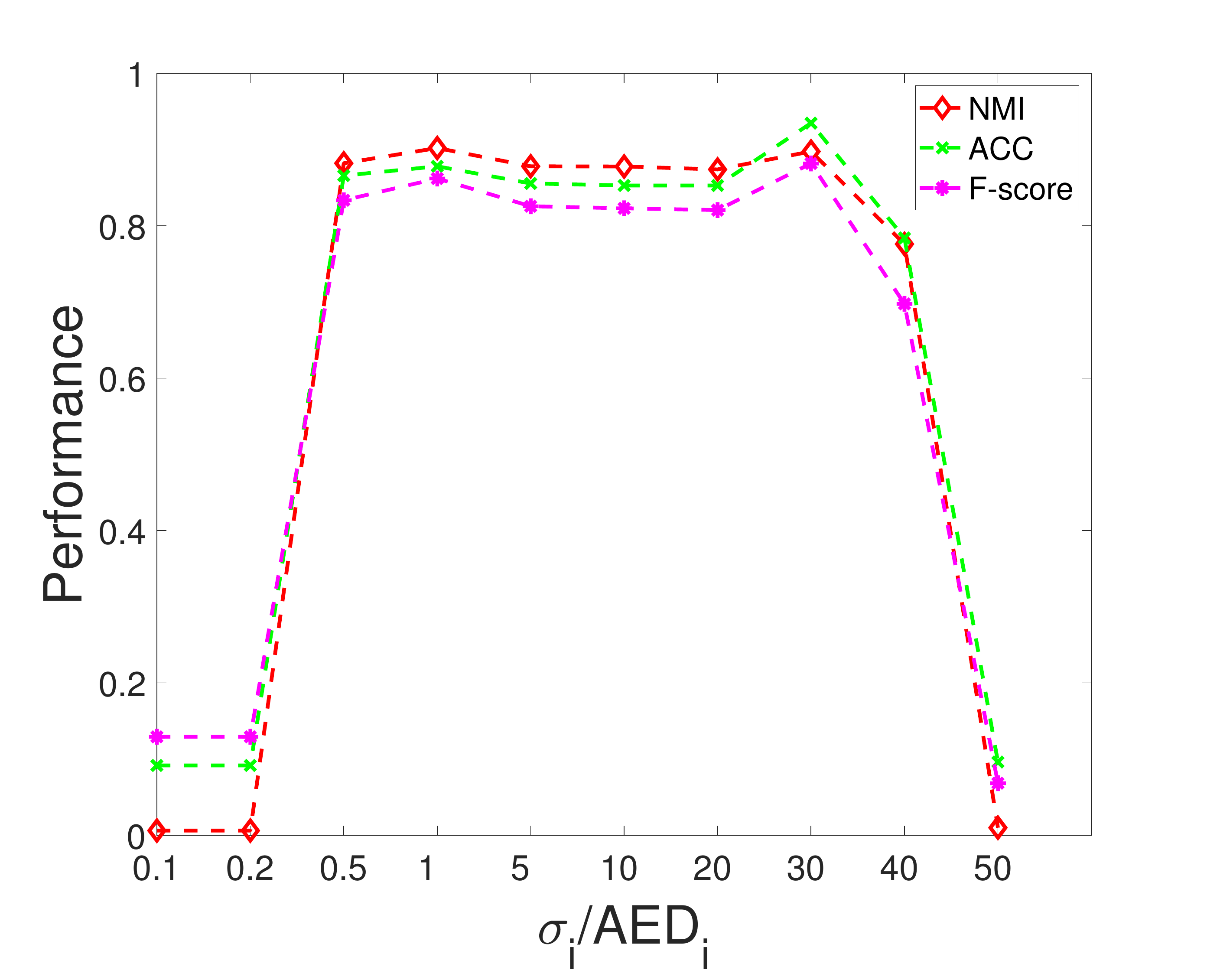}}
	\hspace{-2mm} 
	\subfigure[MITIndoor-67]{ 
		\label{fig:sigma_MIT} 
		\includegraphics[width=2.2in]{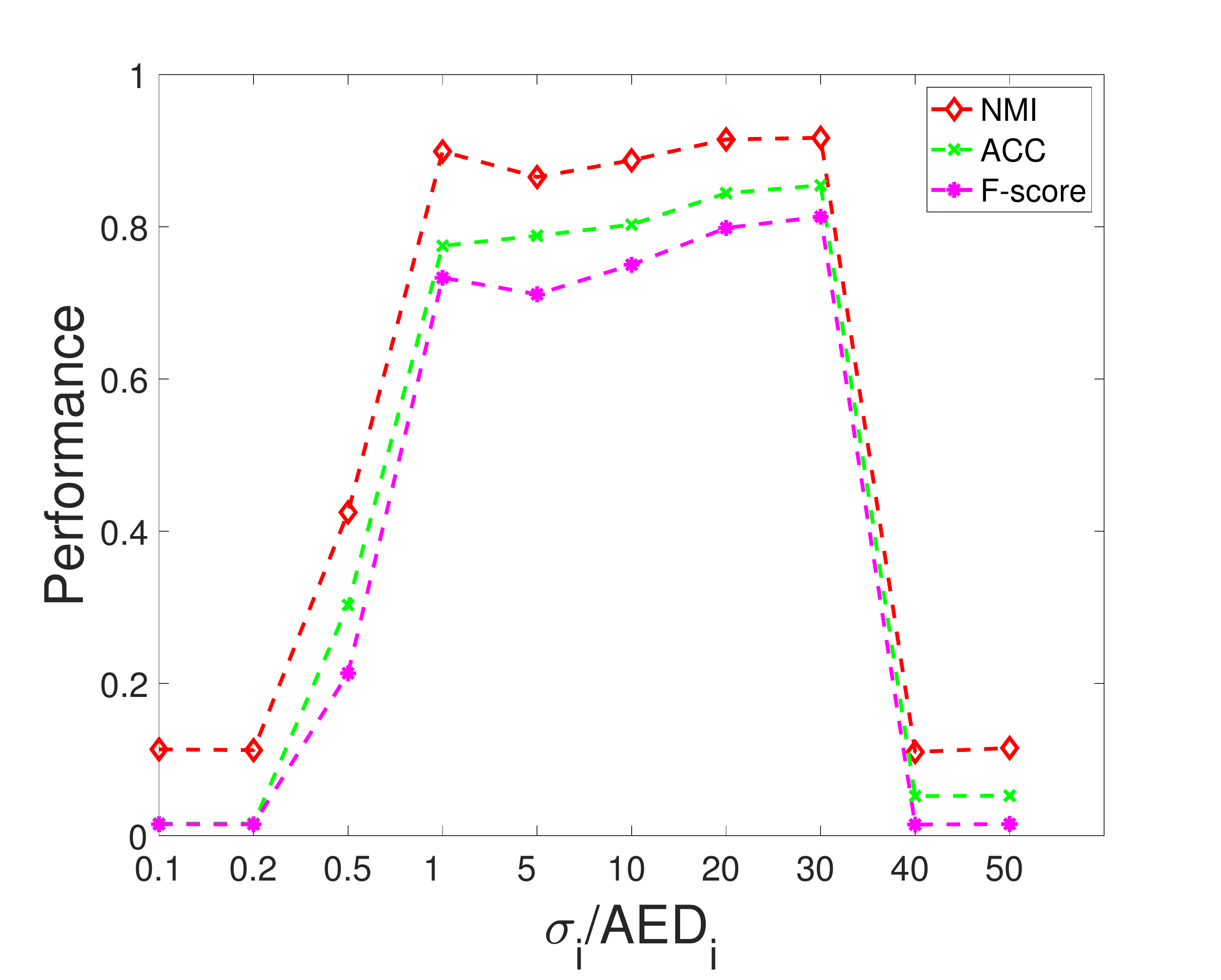}}
	\caption{Influence of $\sigma$ for Gaussian kernel based similarity on the first six datasets.} 
	\label{fig:sigma} 
\end{figure*}
Compared with RMSC, which is also a Markov chain based method, our proposed ETLMSC gains significant improvement.
The main reason is that RMSC only captures the shared information among different view,
while ETLMSC incorporates view-specific information that is useful for clustering.
Based on the t-SVD based tensor nuclear norm to regularize the essential tensor, our method can well preserve these principle components among multi-view representations.
\par
Tensor rotation plays an important role in our methods. Besides the complexity reduction, it can also largely improve the performance, which has already been validated by t-SVD-MSC~\cite{xie2016unifying}. We can see that ETLMSC achieves much better results than UR-ETLMSC  on all datasets. The main reason is that after rotation, we can throughly investigate the complementary information among different views as the SVD is performed on each matrix composed of different view features after FFT.
However, without rotation, the arrangement
of similarity coefficients could be destroyed in Fourier domain, so that complementary information cannot be effectively explored. Therefore, UR-ETLMSC only sometime shows comparable performance with the state-of-the-art methods.
\subsubsection{Parameter Sensitivity Analysis}
There are mainly two parameters in our model, including the balance parameter $\lambda$ and the standard deviation $\sigma$ of Gaussian kernel to compute the similarity.
In experiments,
we find the optimal value for $\lambda$ by grid searching. 
As for $\sigma_{i}$ for the $i$-th view, we directly set it to the average
Euclidean distance~($AED_{i}$) between all $i$-th view features, which is same to RMSC.
We present the evaluation results of our proposed ETLMSC method on the first six datasets with respect to different $\lambda$ and ratio of $\sigma_{i}/AED_i$ in Figs.~\ref{fig:lambda} and ~\ref{fig:sigma}, respectively.
From Fig.~\ref{fig:lambda}, we can observe that on these datasets, the performance of our proposed ETLMSC is relatively stable when $\lambda$ varies in the range of $[0.0008, 0.01]$.
$\lambda$ plays an important role in balancing the contributions of these two parts. When it is very small~(close to $0$), the $\ell_{2,1}$ norm regularization on $\boldsymbol{\mathcal{E}}$ will not work.
$\Vert \boldsymbol{\mathcal{Z}} \Vert_{\circledast}$ will be minimized as much as possible, which leads to $rank(\mathbf{Z}^{(i)})\leq 1$. So the result is very bad.
Moreover, the optimal parameter for each dataset is reported in their corresponding table.
\par
As for $\sigma$, all results of ETLMSC presented in Tables~\ref{tab:bbc-uci}-\ref{tab:caltech} are based on the ratio $\sigma_{i}/AED_{i}=1$.
From Fig.~\ref{fig:sigma}, we can see that our method is not sensitive to this parameter when it varies in a certain large range.  
$\sigma$ controls the discrimination of similarity.
When $\sigma$ is too small~(or too large), all similarities will be close to $0$~(or $1$). It will be hard to distinguish the difference, which leads to bad results.
For all the results reported in the manuscript, they are achieved with $\sigma_{i}/AED_{i}=1$. We can see that with proper ratio, the performance can be further improved, especially on the BBCSport, UCI-Digit, COIL-20, Scene-15, and MITIndoor-67 datasets.
\par
For the parameters $\mu$ and $\rho$ of ADMM, we directly adopt the suggestion of~\cite{LADM} and fix them as $10^{-5}$ and $1.9$, respectively.
These two parameters mainly influence the number of iteration for convergence.
\begin{figure*}[t]
	\centering 
	\subfigure[COIL-20]{ 
		\label{fig:COIL20_error} 
		\includegraphics[width=2.3in]{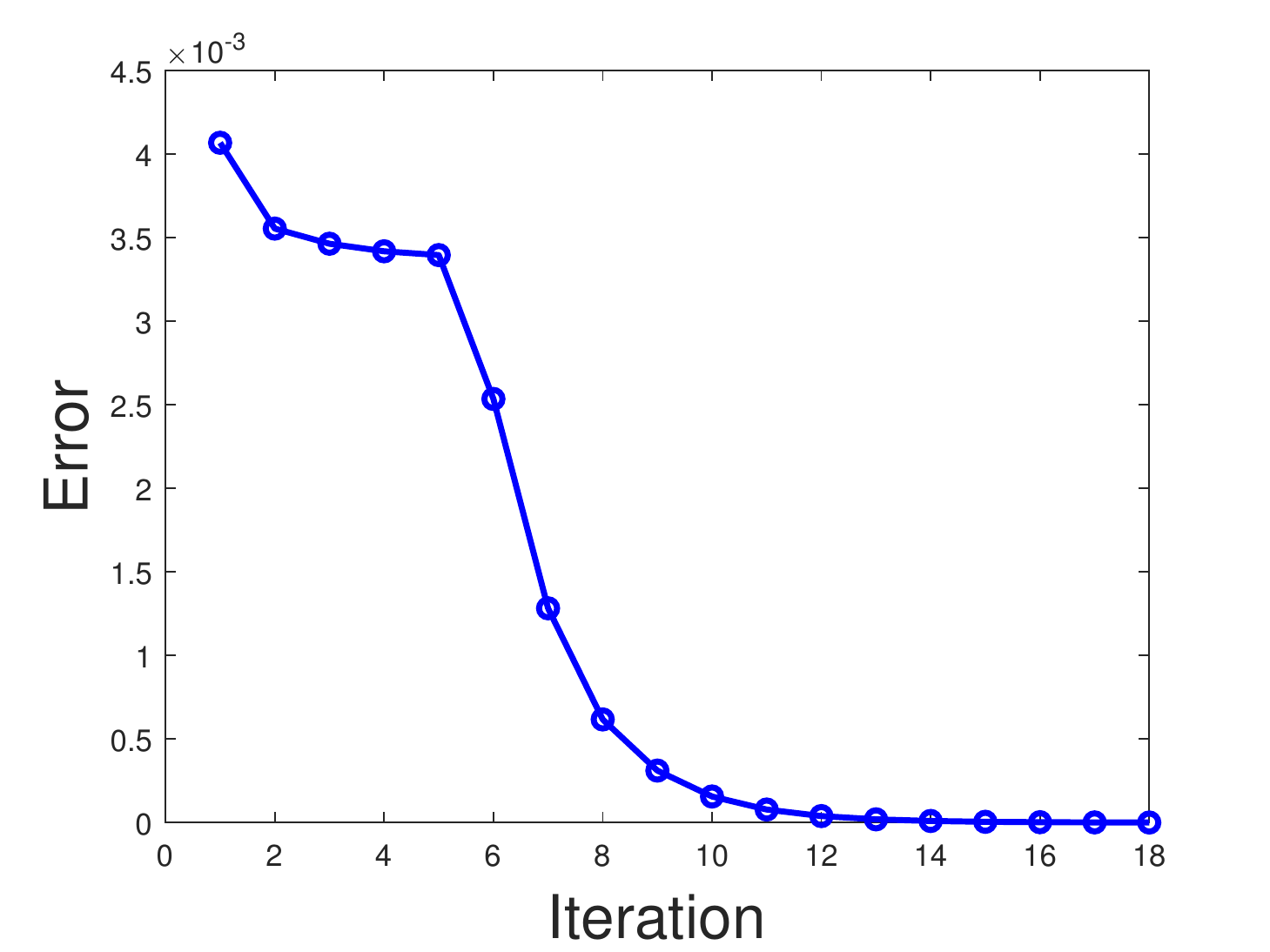}} 
	\hspace{-6mm} 
	\subfigure[Notting-Hill]{ 
		{\label{fig:NH_error}} 
		\includegraphics[width=2.3in]{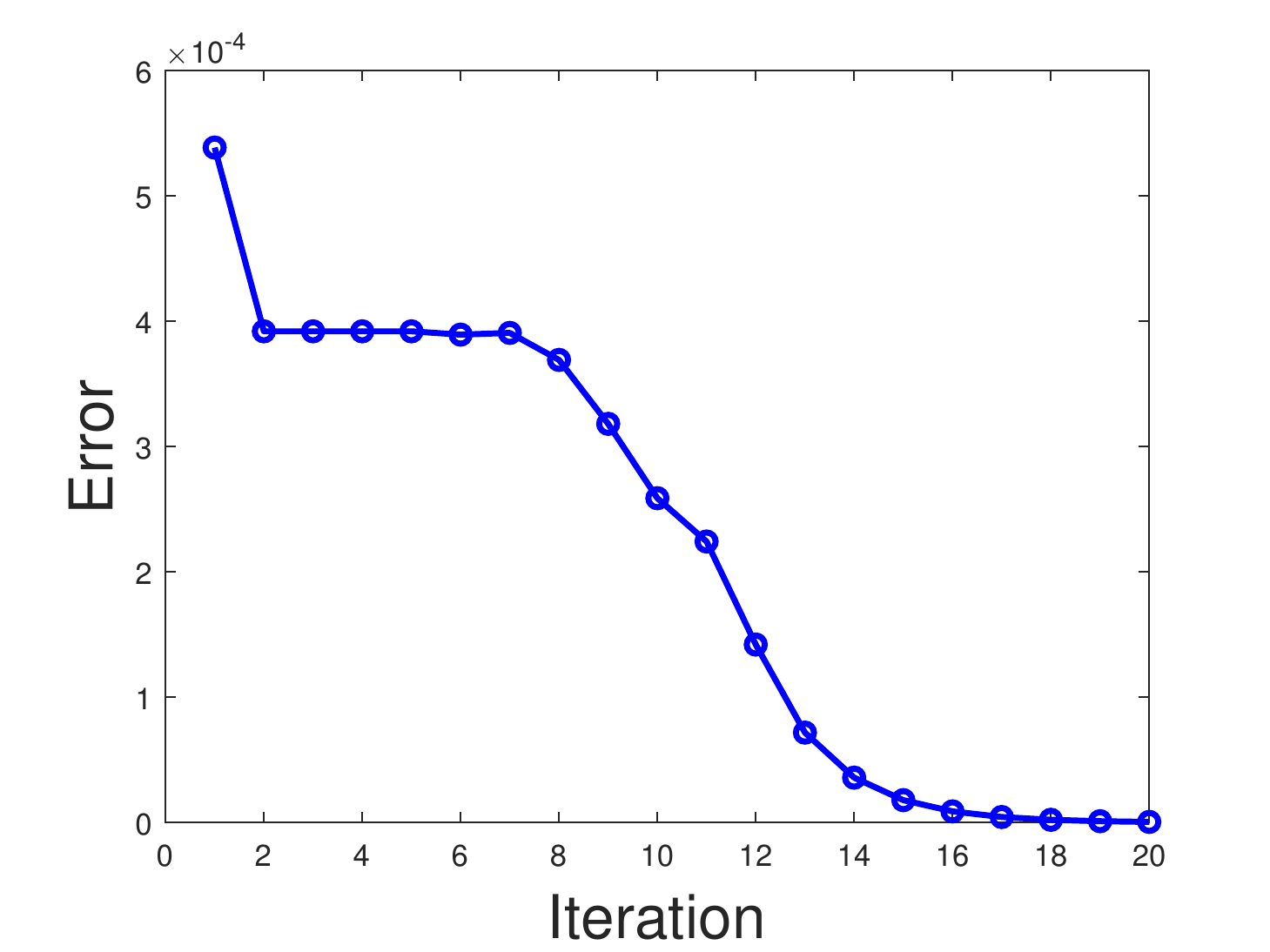}} 
	\hspace{-6mm} 
	\subfigure[MITIndoor-67]{ 
		{\label{fig:MIT}} 
		\includegraphics[width=2.3in]{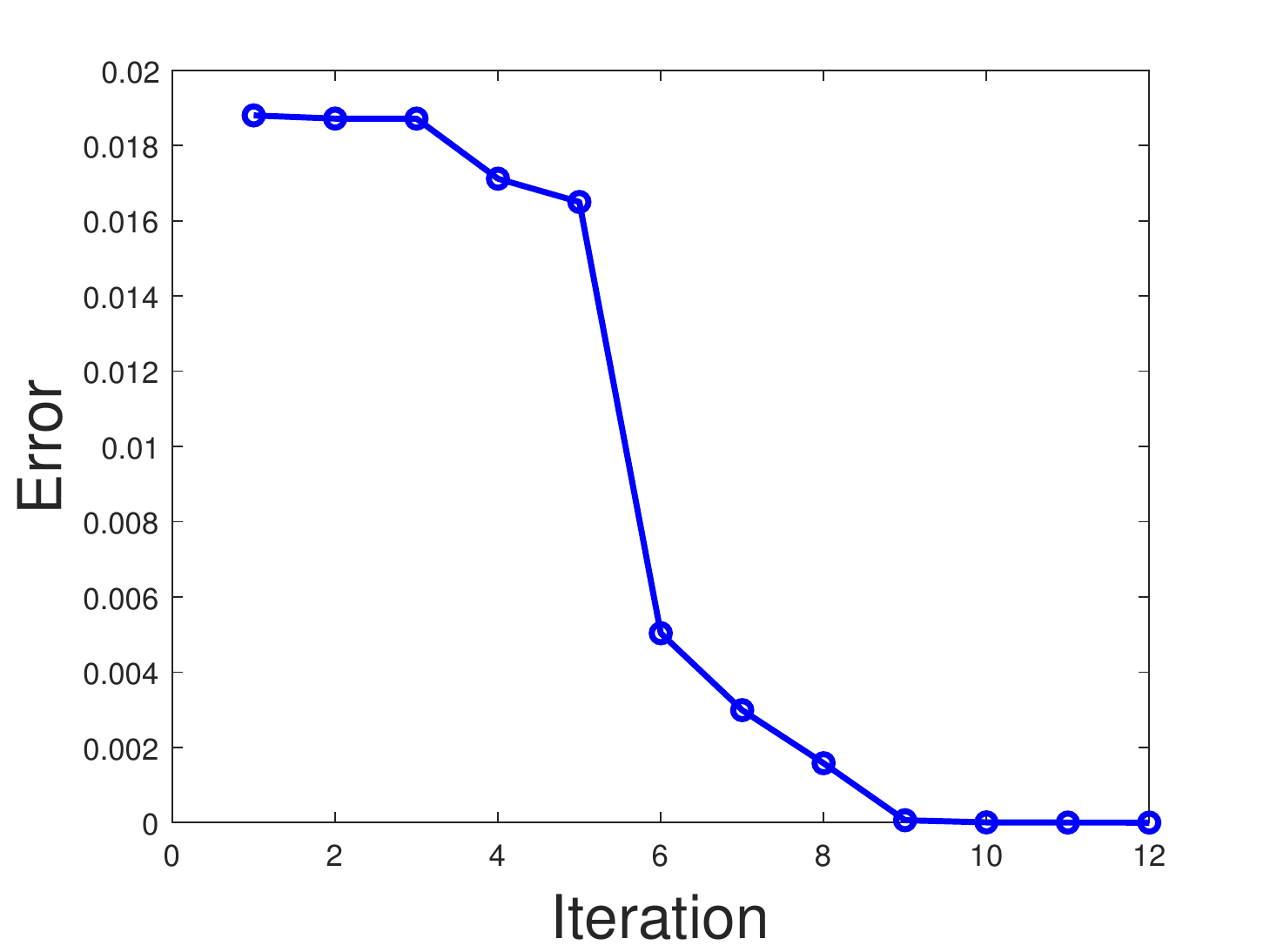}} 
	\caption{Convergence results on the COIL-20, Notting-Hill, and MITIndoor-67 datasets.} 
	\label{fig:error} 
\end{figure*}
\subsubsection{Convergence Analysis}
The theoretical convergence of our algorithm has already been proved in~\cite{LADM}.
In Fig.~\ref{fig:error}, we show the total error of our algorithm in each iteration on the COIL-20, Notting-Hill, and Caltech-101 datasets. 
Here, the total error is defined as the maximum value of changes in each iteration $\Vert \boldsymbol{\mathcal{Z}}^{k+1} - \boldsymbol{\mathcal{Z}}^{k}\Vert_{\infty}$, $\Vert \boldsymbol{\mathcal{E}}^{k+1} - \boldsymbol{\mathcal{E}}^{k}\Vert_{\infty}$, and reconstruction error $\Vert \boldsymbol{\mathcal{\tilde{P}}}-\boldsymbol{\mathcal{Z}}^{k+1} - \boldsymbol{\mathcal{E}}^{k+1}\Vert_{\infty}$:
\begin{equation}
	\text{Error}=\max(
	\Vert \Delta \boldsymbol{\mathcal{Z}} \Vert_{\infty}, 
	\Vert \Delta \boldsymbol{\mathcal{E}} \Vert_{\infty}, 
	\Vert \boldsymbol{\mathcal{\tilde{P}}}-\boldsymbol{\mathcal{Z}}^{k+1} - \boldsymbol{\mathcal{E}}^{k+1}\Vert_{\infty} ).
	\notag
\end{equation}
\begin{table*}[tp]
	\centering
	\renewcommand\arraystretch{1.2}
	\caption{Time complexity and running time to compute affinity matrix on these   datasets of different methods. $K,M,N$ are the number of iterations, views, and samples, respectively. All the time are measured by seconds.}
	\label{tab:time}
	\begin{tabular}{|c|c|c|c|c|c|c|}
		\hline
		Methods     & RMSC & DiMSC & LTMSC & ECMSC & t-SVD-MSC & ETLMSC(Ours) 
		\\ \hline
		Complexity   & $\mathcal{O}(KN^{3})$     & $\mathcal{O}(KMN^{3})$     &  $\mathcal{O}(KMN^{3})$  &  $\mathcal{O}((K+M)N^{3})$      & $\mathcal{O}(MN^{3}+KMN^{2}\log(N))$          & $\mathcal{O}(KMN^{2}\log(N))$       
		\\ \hline
		Time on BBC-Sport         & 4.5     & 35.8     &   23.4    & 78.7    & 10.6    &  \bf{2.1}      
		\\ \hline
		Time on COIL-20         & 74.8     & 1075.1       &  375.9     & 954.2     & 103.4      &  \bf{19.6}      
		\\ \hline
		Time on UCI-Digit         &  214.6    &  2706.4    &  959.3     &  468.5   & 225.7    &  \bf{54.6}      
		\\ \hline
		Time on Notting-Hill         &  2531.3    &  43813.6    &  10408.7     &  6319.3   & 3373.3    &  \bf{562.8}      
		\\ \hline
		Time on Scene-15         &  2407.9    &  38904.7    &   9270.6    &  5663.9   & 2627.8    &  \bf{489.7}      
		\\ \hline
		Time on MITIndoor-67         & 3796.5 &    66274.3      &  15759.2     & 9673.2      &  5957.5     &  \bf{ 930.5}      
		\\ \hline
		Time on Caltech-101         &15710.9      & 218825.5      & 76833.2     &   41558.6    &  18929.7     &  \bf{5395.7}      
		\\ \hline
	\end{tabular}
\end{table*}
According to Fig.~\ref{fig:error}, we can see that the error decreases with the increasing of iteration number.
Our algorithm converges within $20$ iterations, which is also true on other datasets.
As we can compute the close-form solution in each iteration with relatively low computation complexity, our algorithm is very efficient.
\subsubsection{Complexity Comparison}
In Table~\ref{tab:time}, we present computation complexity and running time of the state-of-the-art methods on all these datasets.
Since all these methods share the similar post-processing procedure that has the same complexity, we only report the computational complexity and running time for learning the affinity matrix.
We need to mention that the number of iteration $K$ has an obvious affect on the running time,
and parameter selection will influence $K$. 
So we can see that the running time of ECMSC on the UCI-Digit dataset could be shorter than that on the COIL-20 dataset.
We can see that our method has the lowest complexity and  the  shortest  processing  time among these related approaches on all datasets, which demonstrates the efficiency of our proposed method. 
For example, 
on the COIL-20 dataset, our algorithm can finish within $20$ seconds, while the second best method RMSC needs more than $70$ seconds, and t-SVD-MSC costs more than $100$ seconds.
On the largest Caltech-101 dataset, our method can save much time compared with t-SVD-MSC.

\begin{figure*}[!ht]
	\centering 
	\subfigure[RMSC]{ 
		{\label{fig:vis_rmsc}} 
		\includegraphics[width=3.5in]{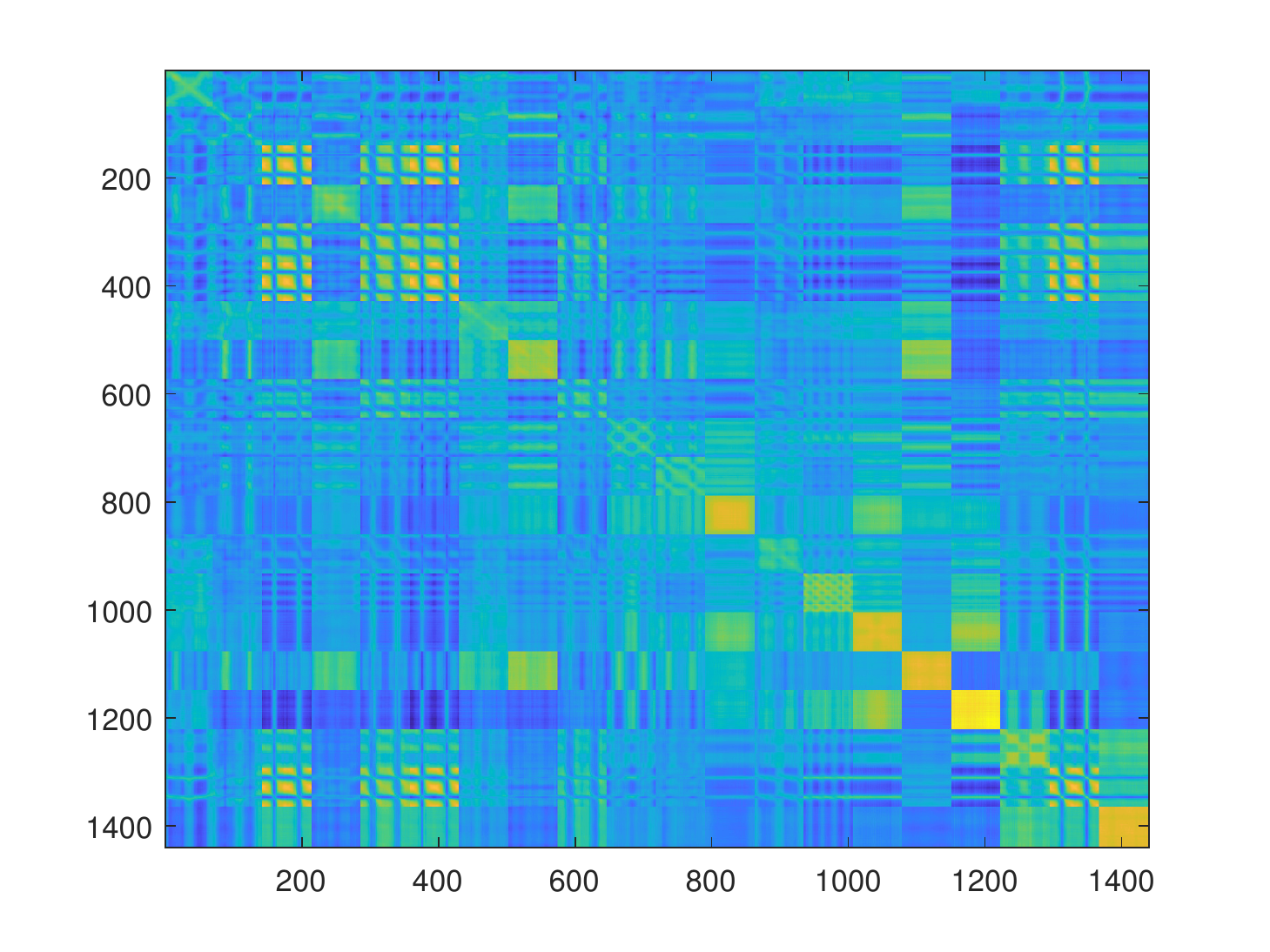}} 
	\hspace{-6mm} 
	\subfigure[ETLMSC]{ 
		\label{fig:vis_etlmsc} 
		\includegraphics[width=3.52in]{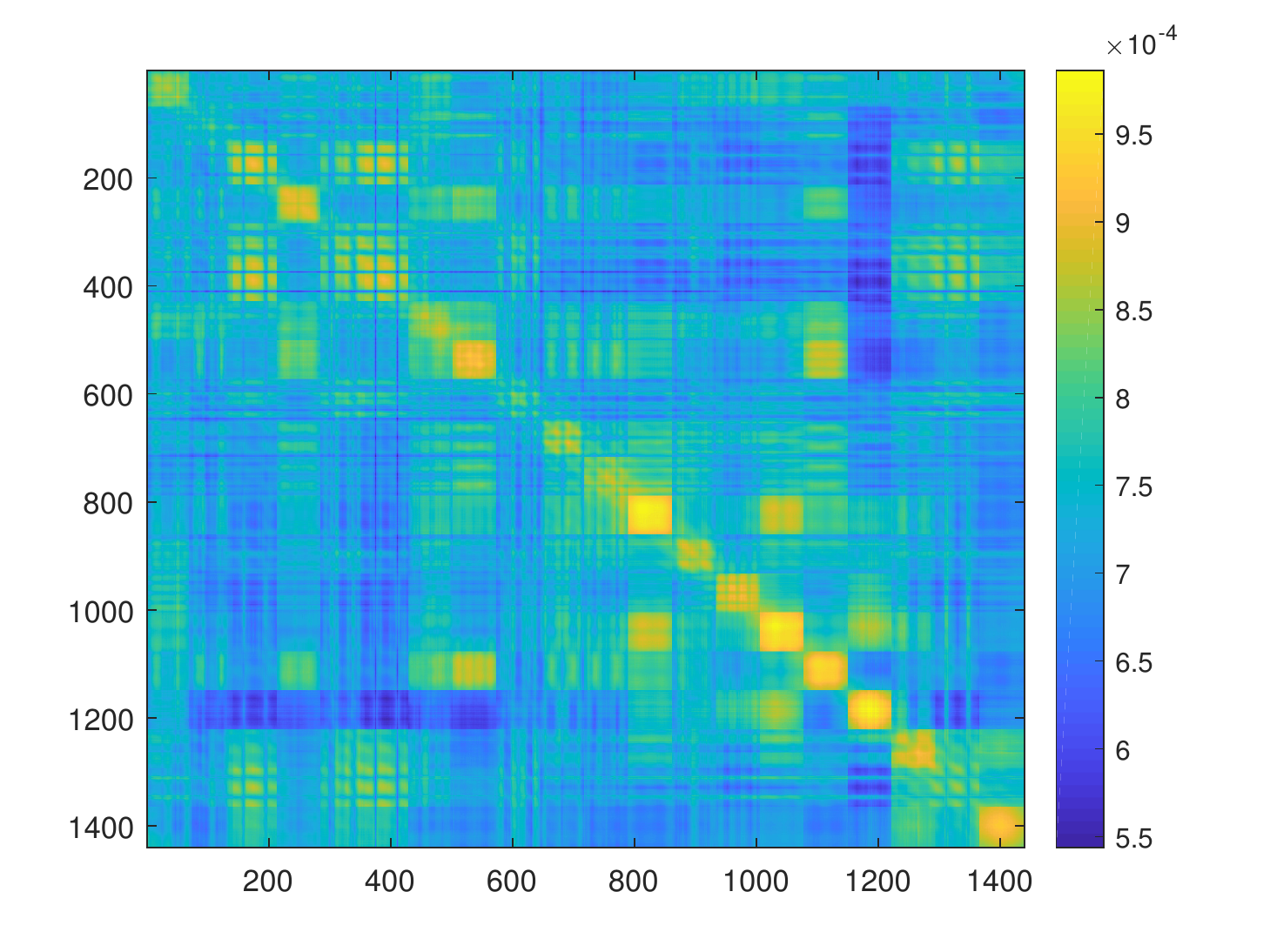}} 
	\caption{Visualization of learned transition probability matrices of two spectral clustering based methods on the COIL-20 dataset.} 
	\label{fig:vis_coil} 
\end{figure*}
\subsubsection{Representation Visualization}
In Fig.~\ref{fig:vis_coil}, we show the visualization of the learned optimal transition probability matrix.
Due to the limition of space, we only present the results of two Markov chain based spectral clustering methods~(RMSC and our proposed ETLMSC) on the COIL-20 dataset.
For ETLMSC, the transition probability matrix is computed by the average of lateral slices of the optimal essential tensor $\boldsymbol{\mathcal{Z}}$.
The yellow color represents the large value.
Compared with the result of RMSC in Fig.~\ref{fig:vis_rmsc}, we can easily see that the result of ETLMSC in Fig.~\ref{fig:vis_etlmsc} is much better as most large values concentrate on the diagonal blocks.
This can also be verified by comparing the experimental results in Tables~\ref{tab:bbc-uci}-\ref{tab:scene-mit}.
While RMSC only captures shared information among different views,
it is more meaningful for our ETLMSC method to explore high order multi-view correlations based on tensor formulation. 
\subsubsection{Comparison with t-SVD-MSC}
t-SVD-MSC~\cite{xie2016unifying} achieves very good performance for the task of multi-view clustering.
Both the proposed ETLMSC and t-SVD-MSC~\cite{xie2016unifying} are based on the tensor nuclear norm defined by the t-SVD for multi-view clustering.
But there are many differences.
First, construction of affinity matrix and tensor is totally different. We adopt the Markov chain to compute the transition probability matrix, while t-SVD-MSC is based on self-representation, which is of high computation complexity and under the assumption of subspace structure. 
Second, the model and optimization process are much different. 
We directly divide the transition probability tensor into two parts with low-rank and sparse constraints, while their method need to optimize the self-representation coefficients. So the optimization process is also different.
Most importantly, compared with t-SVD-MSC, based on the experimental results presented above, our method achieves better performance with much lower complexity and less processing time.
\section{Conclusion and future work} 
In this paper, we propose a novel essential tensor learning method
for Markov chain based multi-view spectral clustering.
Based on multi-view transition probability matrices, we construct a $3$-order tensor.
We explore the high order correlations among multiple views by learning the essential tensor with low-rank constraint based on t-SVD based tensor nuclear norm.
With tensor rotation operation,
the proposed algorithm can be optimized efficiently and the principle components can be well preserved.
We evaluate the performance of our method on seven datasets with respect to different applications,
and it achieves superior performance compared with the state-of-the-art methods.
\par
For future work, we would like to focus on the fast and scalable algorithms, such as the sampling technique or recover the subspace of the whole tensor with a much smaller seed tensor. So that the computation complexity of the proposed model can be further reduced, which will make ETLMSC much suitable for large-scale applications.

%


\ifCLASSOPTIONcompsoc
  \section*{Acknowledgments}
\else
  \section*{Acknowledgment}
\fi

We would like to thank Dr. Yuan Xie for his selfless support in sharing codes and datasets as well as the valuable suggestions.

\ifCLASSOPTIONcaptionsoff
  \newpage
\fi



\bibliography{egbib}

\begin{thebibliography}{10}

\bibitem{ng2002spectral}
A.~Y. Ng, M.~I. Jordan, and Y.~Weiss, ``On spectral clustering: Analysis and an
  algorithm,'' in {\em Proceedings of the Neural Information Processing
  Systems}, pp.~849--856, 2002.

\bibitem{elhamifar2013sparse}
E.~Elhamifar and R.~Vidal, ``Sparse subspace clustering: Algorithm, theory, and
  applications,'' {\em {IEEE} Transactions on Pattern Analysis and Machine
  Intelligence}, vol.~35, no.~11, pp.~2765--2781, 2013.

\bibitem{liu2013robust}
G.~Liu, Z.~Lin, S.~Yan, J.~Sun, Y.~Yu, and Y.~Ma, ``Robust recovery of subspace
  structures by low-rank representation,'' {\em {IEEE} Transactions on Pattern
  Analysis and Machine Intelligence}, vol.~35, no.~1, pp.~171--184, 2013.

\bibitem{kumar2011coreg}
A.~Kumar, P.~Rai, and H.~Daume, ``Co-regularized multi-view spectral
  clustering,'' in {\em Proceedings of the Neural Information Processing
  Systems}, pp.~1413--1421, 2011.

\bibitem{cao2015diversity}
X.~Cao, C.~Zhang, H.~Fu, S.~Liu, and H.~Zhang, ``Diversity-induced multi-view
  subspace clustering,'' in {\em Proceedings of the {IEEE} International
  Conference on Computer Vision and Pattern Recognition}, pp.~586--594, 2015.

\bibitem{RMSC}
R.~Xia, Y.~Pan, L.~Du, and J.~Yin, ``Robust multi-view spectral clustering via
  low-rank and sparse decomposition.,'' in {\em Proceedings of the AAAI},
  pp.~2149--2155, 2014.

\bibitem{wang2017exclusivity}
X.~Wang, X.~Guo, Z.~Lei, C.~Zhang, and S.~Z. Li, ``Exclusivity-consistency
  regularized multi-view subspace clustering,'' in {\em Proceedings of the
  {IEEE} International Conference on Computer Vision and Pattern Recognition},
  pp.~923--931, 2017.

\bibitem{xie2018implicit}
X.~Xie, X.~Guo, G.~Liu, and J.~Wang, ``Implicit block diagonal low-rank
  representation,'' {\em IEEE Transactions on Image Processing}, vol.~27,
  no.~1, pp.~477--489, 2018.

\bibitem{liu2010robust}
G.~Liu, Z.~Lin, and Y.~Yu, ``Robust subspace segmentation by low-rank
  representation,'' in {\em Proceedings of the International Conference on
  Machine Learning}, pp.~663--670, 2010.

\bibitem{RPCA}
J.~Wright, A.~Ganesh, S.~Rao, Y.~Peng, and Y.~Ma, ``Robust principal component
  analysis: Exact recovery of corrupted low-rank matrices via convex
  optimization,'' in {\em Proceedings of the Neural Information Processing
  Systems}, pp.~2080--2088, 2009.

\bibitem{TRPCA}
C.~Lu, J.~Feng, Y.~Chen, W.~Liu, Z.~Lin, and S.~Yan, ``Tensor robust principal
  component analysis: Exact recovery of corrupted low-rank tensors via convex
  optimization,'' in {\em Proceedings of the {IEEE} International Conference on
  Computer Vision and Pattern Recognition}, pp.~5249--5257, 2016.

\bibitem{RTPCA}
P.~Zhou and J.~Feng, ``Outlier-robust tensor pca,'' in {\em Proceedings of the
  {IEEE} International Conference on Computer Vision and Pattern Recognition},
  pp.~3938--3946, 2017.

\bibitem{zhou2018tensor}
P.~Zhou, C.~Lu, Z.~Lin, and C.~Zhang, ``Tensor factorization for low-rank
  tensor completion,'' {\em {IEEE} Transactions on Image Processing}, vol.~27,
  no.~3, pp.~1152--1163, 2018.

\bibitem{kong2018t}
H.~Kong, X.~Xie, and Z.~Lin, ``t-schatten-$p$ norm for low-rank tensor
  recovery,'' {\em IEEE Journal of Selected Topics in Signal Processing}, 2018.

\bibitem{carroll1970analysis}
J.~D. Carroll and J.-J. Chang, ``Analysis of individual differences in
  multidimensional scaling via an n-way generalization of “eckart-young”
  decomposition,'' {\em Psychometrika}, vol.~35, no.~3, pp.~283--319, 1970.

\bibitem{harshman1970foundations}
R.~A. Harshman, ``Foundations of the parafac procedure: Models and conditions
  for an" explanatory" multimodal factor analysis,'' 1970.

\bibitem{tucker1966some}
L.~R. Tucker, ``Some mathematical notes on three-mode factor analysis,'' {\em
  Psychometrika}, vol.~31, no.~3, pp.~279--311, 1966.

\bibitem{Tensor}
M.~E. Kilmer, K.~Braman, N.~Hao, and R.~C. Hoover, ``Third-order tensors as
  operators on matrices: A theoretical and computational framework with
  applications in imaging,'' {\em {SIAM} Journal on Matrix Analysis and
  Applications}, vol.~34, no.~1, pp.~148--172, 2013.

\bibitem{huang2014provable}
B.~Huang, C.~Mu, D.~Goldfarb, and J.~Wright, ``Provable low-rank tensor
  recovery,'' {\em Optimization-Online}, vol.~4252, p.~2, 2014.

\bibitem{zhang2014novel}
Z.~Zhang, G.~Ely, S.~Aeron, N.~Hao, and M.~Kilmer, ``Novel methods for
  multilinear data completion and de-noising based on tensor-svd,'' in {\em
  Proceedings of the {IEEE} International Conference on Computer Vision and
  Pattern Recognition}, pp.~3842--3849, 2014.

\bibitem{kumar2011cotrain}
A.~Kumar and H.~Daum{\'e}, ``A co-training approach for multi-view spectral
  clustering,'' in {\em Proceedings of the International Conference on Machine
  Learning}, pp.~393--400, 2011.

\bibitem{wang2014multi}
H.~Wang, C.~Weng, and J.~Yuan, ``Multi-feature spectral clustering with minimax
  optimization,'' in {\em Proceedings of the {IEEE} International Conference on
  Computer Vision and Pattern Recognition}, pp.~4106--4113, 2014.

\bibitem{shi2000normalized}
J.~Shi and J.~Malik, ``Normalized cuts and image segmentation,'' {\em {IEEE}
  Transactions on Pattern Analysis and Machine Intelligence}, vol.~22, no.~8,
  pp.~888--905, 2000.

\bibitem{zhou2005learning}
D.~Zhou, J.~Huang, and B.~Sch{\"o}lkopf, ``Learning from labeled and unlabeled
  data on a directed graph,'' in {\em Proceedings of the International
  Conference on Machine Learning}, pp.~1036--1043, 2005.

\bibitem{zhou2007spectral}
D.~Zhou and C.~J. Burges, ``Spectral clustering and transductive learning with
  multiple views,'' in {\em Proceedings of the International Conference on
  Machine Learning}, pp.~1159--1166, 2007.

\bibitem{wang2018multiview}
Y.~Wang, L.~Wu, X.~Lin, and J.~Gao, ``Multiview spectral clustering via
  structured low-rank matrix factorization,'' {\em IEEE Transactions on Neural
  Networks and Learning Systems}, 2018.

\bibitem{brbic2018multi}
M.~Brbi{\'c} and I.~Kopriva, ``Multi-view low-rank sparse subspace
  clustering,'' {\em Pattern Recognition}, vol.~73, pp.~247--258, 2018.

\bibitem{zhang2015low}
C.~Zhang, H.~Fu, S.~Liu, G.~Liu, and X.~Cao, ``Low-rank tensor constrained
  multiview subspace clustering,'' in {\em Proceedings of the {IEEE}
  International Conference on Computer Vision}, pp.~1582--1590, 2015.

\bibitem{xie2016unifying}
Y.~Xie, D.~Tao, W.~Zhang, Y.~Liu, L.~Zhang, and Y.~Qu, ``On unifying multi-view
  self-representations for clustering by tensor multi-rank minimization,'' {\em
  International Journal of Computer Vision}, pp.~1--23, 2018.

\bibitem{yin2018multiview}
M.~Yin, J.~Gao, S.~Xie, and Y.~Guo, ``Multiview subspace clustering via
  tensorial t-product representation,'' {\em IEEE Transactions on Neural
  Networks and Learning Systems}, no.~99, pp.~1--14, 2018.

\bibitem{zhang2017latent}
C.~Zhang, Q.~Hu, H.~Fu, P.~Zhu, and X.~Cao, ``Latent multi-view subspace
  clustering,'' in {\em Proceedings of the {IEEE} International Conference on
  Computer Vision and Pattern Recognition}, vol.~30, pp.~4279--4287, 2017.

\bibitem{zhang2018generalized}
C.~Zhang, H.~Fu, Q.~Hu, X.~Cao, Y.~Xie, D.~Tao, and D.~Xu, ``Generalized latent
  multi-view subspace clustering,'' {\em IEEE Transactions on Pattern Analysis
  and Machine Intelligence}, 2018.

\bibitem{li2015structured}
C.-G. Li and R.~Vidal, ``Structured sparse subspace clustering: A unified
  optimization framework,'' in {\em Proceedings of the {IEEE} International
  Conference on Computer Vision and Pattern Recognition}, pp.~277--286, 2015.

\bibitem{chaudhuri2009multi}
K.~Chaudhuri, S.~M. Kakade, K.~Livescu, and K.~Sridharan, ``Multi-view
  clustering via canonical correlation analysis,'' in {\em Proceedings of the
  International Conference on Machine Learning}, pp.~129--136, 2009.

\bibitem{cortes2009learning}
C.~Cortes, M.~Mohri, and A.~Rostamizadeh, ``Learning non-linear combinations of
  kernels,'' in {\em Proceedings of the Neural Information Processing Systems},
  pp.~396--404, 2009.

\bibitem{xu2016discriminatively}
J.~Xu, J.~Han, and F.~Nie, ``Discriminatively embedded k-means for multi-view
  clustering,'' in {\em Proceedings of the {IEEE} International Conference on
  Computer Vision and Pattern Recognition}, pp.~5356--5364, 2016.

\bibitem{kmeans}
J.~A. Hartigan and M.~A. Wong, ``Algorithm as 136: A k-means clustering
  algorithm,'' {\em Journal of the Royal Statistical Society. Series C (Applied
  Statistics)}, vol.~28, no.~1, pp.~100--108, 1979.

\bibitem{LADM}
Z.~Lin, R.~Liu, and Z.~Su, ``Linearized alternating direction method with
  adaptive penalty for low-rank representation,'' in {\em Proceedings of the
  Neural Information Processing Systems}, pp.~612--620, 2011.

\bibitem{TWIST}
W.~Hu, D.~Tao, W.~Zhang, Y.~Xie, and Y.~Yang, ``The twist tensor nuclear norm
  for video completion,'' {\em {IEEE} Transactions on Nuural Networks and
  Learning Systems}, pp.~1--13, 2017.

\bibitem{greene06icml}
D.~Greene and P.~Cunningham, ``Practical solutions to the problem of diagonal
  dominance in kernel document clustering,'' in {\em Proceedings of the
  International Conference on Machine Learning}, pp.~377--384, 2006.

\bibitem{UCI}
A.~Asuncion and D.~Newman, ``{UCI} machine learning repository,'' 2007.

\bibitem{LBP}
T.~Ojala, M.~Pietikainen, and T.~Maenpaa, ``Multiresolution gray-scale and
  rotation invariant texture classification with local binary patterns,'' {\em
  {IEEE} Transactions on Pattern Analysis and Machine Intelligence}, vol.~24,
  no.~7, pp.~971--987, 2002.

\bibitem{gabor}
M.~Lades, J.~C. Vorbruggen, J.~Buhmann, J.~Lange, C.~Von Der~Malsburg, R.~P.
  Wurtz, and W.~Konen, ``Distortion invariant object recognition in the dynamic
  link architecture,'' {\em {IEEE} Transactions on Computers}, vol.~42, no.~3,
  pp.~300--311, 1993.

\bibitem{zhang2009character}
Y.-F. Zhang, C.~Xu, H.~Lu, and Y.-M. Huang, ``Character identification in
  feature-length films using global face-name matching,'' {\em {IEEE}
  Transactions on Multimedia}, vol.~11, no.~7, pp.~1276--1288, 2009.

\bibitem{fei2005bayesian}
L.~Fei-Fei and P.~Perona, ``A bayesian hierarchical model for learning natural
  scene categories,'' in {\em Proceedings of the {IEEE} International
  Conference on Computer Vision and Pattern Recognition}, pp.~524--531, 2005.

\bibitem{PHOW}
A.~Bosch, A.~Zisserman, and X.~Munoz, ``Image classification using random
  forests and ferns,'' in {\em Proceedings of the {IEEE} International
  Conference on Computer Vision}, pp.~1--8, 2007.

\bibitem{wu2011centrist}
J.~Wu and J.~M. Rehg, ``Centrist: A visual descriptor for scene
  categorization,'' {\em {IEEE} Transactions on Pattern Analysis and Machine
  Intelligence}, vol.~33, no.~8, pp.~1489--1501, 2011.

\bibitem{quattoni2009recognizing}
A.~Quattoni and A.~Torralba, ``Recognizing indoor scenes,'' in {\em Proceedings
  of the {IEEE} International Conference on Computer Vision and Pattern
  Recognition}, pp.~413--420, 2009.

\bibitem{simonyan2014very}
K.~Simonyan and A.~Zisserman, ``Very deep convolutional networks for
  large-scale image recognition,'' {\em arXiv preprint arXiv:1409.1556}, 2014.

\bibitem{fei2007learning}
L.~Fei-Fei, R.~Fergus, and P.~Perona, ``Learning generative visual models from
  few training examples: An incremental bayesian approach tested on 101 object
  categories,'' {\em Computer Vision and Image Understanding}, vol.~106, no.~1,
  pp.~59--70, 2007.

\bibitem{szegedy2016rethinking}
C.~Szegedy, V.~Vanhoucke, S.~Ioffe, J.~Shlens, and Z.~Wojna, ``Rethinking the
  inception architecture for computer vision,'' in {\em Proceedings of the
  {IEEE} International Conference on Computer Vision and Pattern Recognition},
  pp.~2818--2826, 2016.

\bibitem{arpit2014dimensionality}
D.~Arpit, I.~Nwogu, and V.~Govindaraju, ``Dimensionality reduction with
  subspace structure preservation,'' in {\em Proceedings of the Neural
  Information Processing Systems}, pp.~712--720, 2014.

\bibitem{zhang2014jointly}
G.~Zhang, R.~He, and L.~S. Davis, ``Jointly learning dictionaries and subspace
  structure for video-based face recognition,'' in {\em Proceedings of the
  {IEEE} Asian Conference on Computer Vision}, pp.~97--111, 2014.

\end{thebibliography}


%
%
%

\begin{IEEEbiography}[{\includegraphics[width=1in,height=1.25in,clip,keepaspectratio]{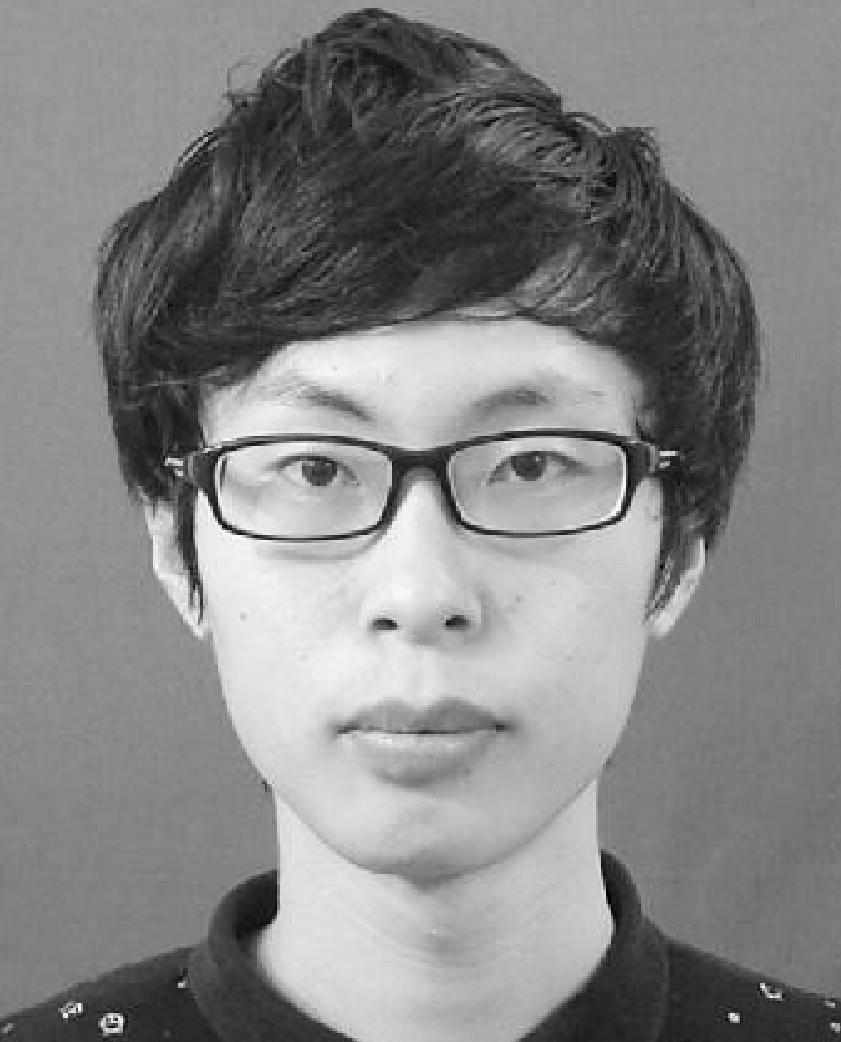}}]{Jianlong Wu}
	received the B.E. degree in electronics and information engineering from Huazhong University of Science and Technology in 2014. He is currently pursuing the Ph.D. degree with the School of Electronics Engineering and Computer Science, Peking University. His research interests include computer vision, pattern recognition and machine learning.
\end{IEEEbiography}
\vspace{-5mm}
\begin{IEEEbiography}[{\includegraphics[width=1in,height=1.25in,clip,keepaspectratio]{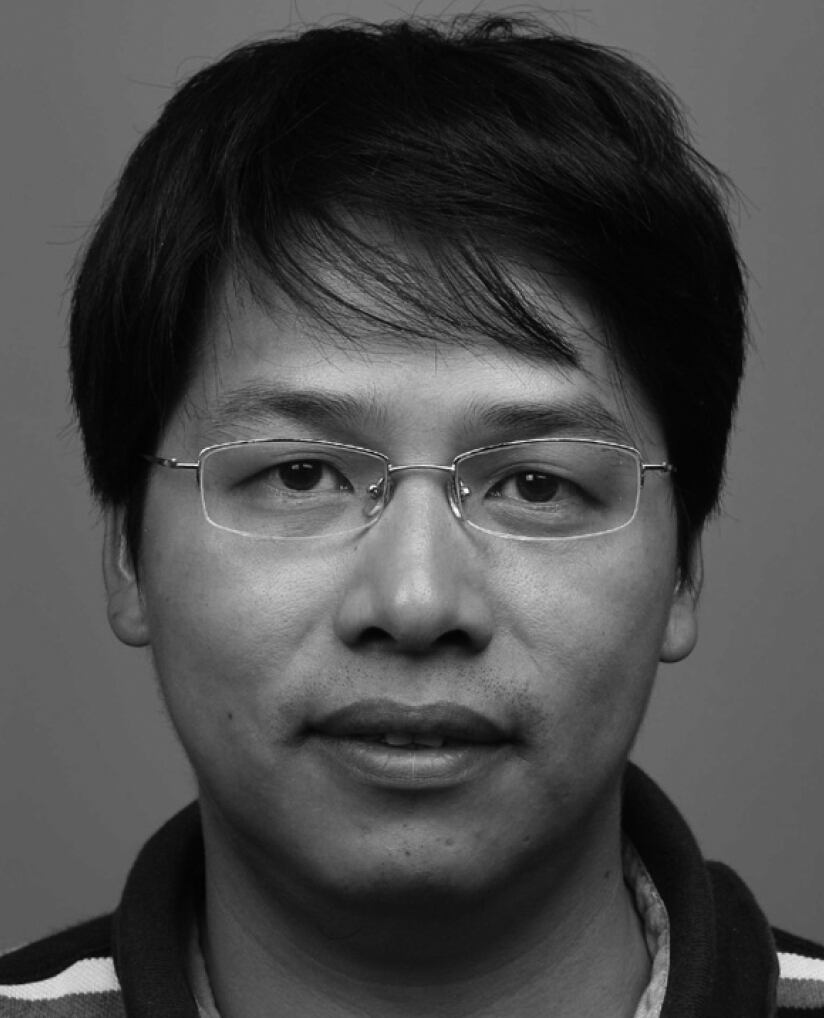}}]{Zhouchen Lin}(M'00-SM'08-F'18)
	received the Ph.D. degree in applied mathematics from Peking University in 2000. He is currently a Professor with the Key Laboratory of Machine Perception (MOE), School of Electronics Engineering and Computer Science, Peking University. His research interests
	include computer vision, image processing, machine learning, pattern recognition, and numerical optimization. He is a fellow of the IAPR and the IEEE. He is an Associate Editor of the IEEE Transactions on Pattern Analysis and Machine Intelligence and the International Journal of Computer Vision, an Area Chair of ACCV 2009/2018, CVPR 2014/2016/2019, ICCV 2015, NIPS 2015/2018/2019, and AAAI 2019, and a Senior Program Committee Member of AAAI 2016/2017/2018 and IJCAI 2016/2018/2019.
\end{IEEEbiography}
\begin{IEEEbiography}[{\includegraphics[width=1in,height=1.25in,clip,keepaspectratio]{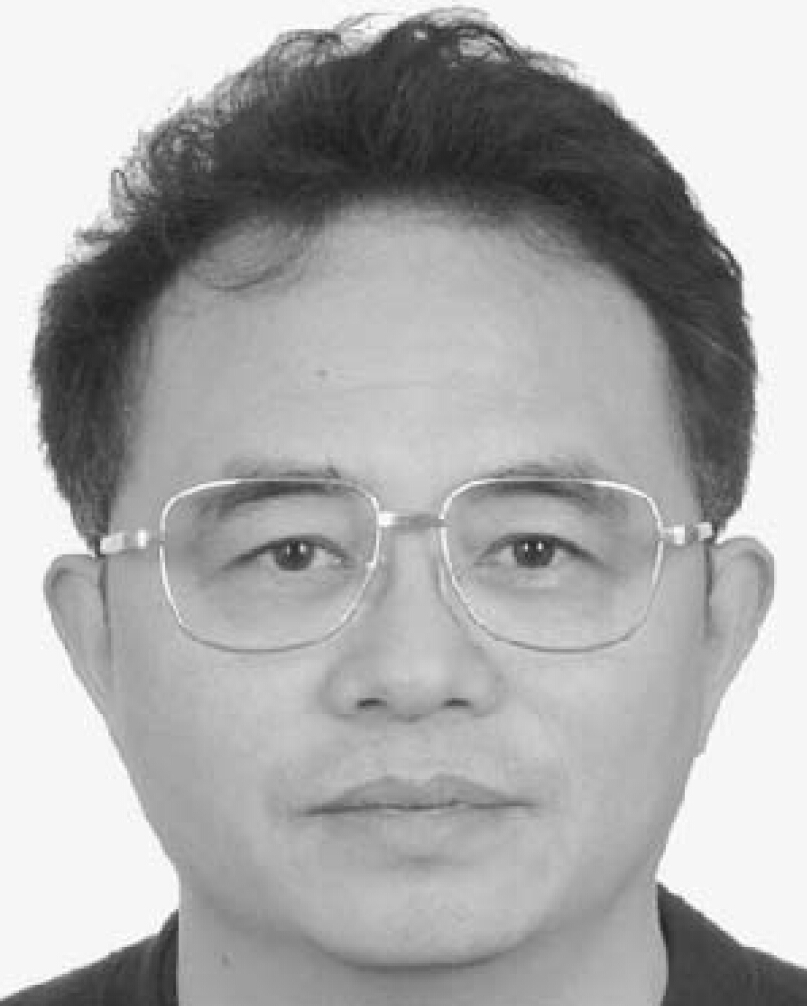}}]{Hongbin Zha}(M'06)
	received the M.S. and Ph.D. degrees in electrical engineering from Kyushu
	University, Fukuoka, Japan, in 1987 and 1990, respectively. He joined Kyushu University in 1991 as an associate professor. He was a Research Associate with the Kyushu Institute of Technology. He was also a Visiting Professor with the Center for Vision,
	Speech, and Signal Processing, Surrey University, U.K., in 1999. Since 2000, he has been a Professor with the Key Laboratory of Machine Perception, Peking University, Beijing, China. He has authored more than 300 technical publications in journals, books, and international conference proceedings. His research interests include computer vision, digital geometry processing, and robotics. He received the Franklin V. Taylor Award
	from the IEEE Systems, Man, and Cybernetics Society in 1999. He is a member of the IEEE Computer Society.
\end{IEEEbiography}

\end{document}